%% file: main.tex
\documentclass[sigconf]{acmart}
\AtBeginDocument{%
  }

\copyrightyear{2026}
\acmYear{2026}
\setcopyright{cc}
\setcctype{by}
\acmConference[MM '26]{Proceedings of the 34th ACM International Conference on Multimedia}{November 10--14, 2026}{Rio de Janeiro, Brazil}
\acmBooktitle{Proceedings of the 34th ACM International Conference on Multimedia (MM '26), November 10--14, 2026, Rio de Janeiro, Brazil}
\acmDOI{10.1145/3767308.3835470}
\acmISBN{979-8-4007-2213-4/2026/11}

\usepackage{enumitem}
\usepackage{colortbl}
\newcommand{\model}{DwD}
\newcommand{\modelfull}{Driving with DINO}
\definecolor{agc}{RGB}{255, 240, 170} 
\definecolor{asc}{RGB}{230, 230, 230} 
\definecolor{abc}{RGB}{255, 218, 185} 

\newcommand{\first}[1]{\cellcolor{agc}#1}
\newcommand{\second}[1]{\cellcolor{asc}#1}
\newcommand{\third}[1]{\cellcolor{abc}#1}

\begin{document}

\title{Driving with DINO: Vision Foundation Features as a Unified Bridge for Sim-to-Real Generation in Autonomous Driving}

\author{Xuyang Chen}
\authornote{These authors contributed equally to this research.}
\affiliation{%
  \institution{Technical University of Munich}
  \city{Munich}
  \country{Germany}
}
\affiliation{%
  \institution{Huawei Hilbert Research Center}
  \city{Munich}
  \country{Germany}
}
\email{xuyang.chen@tum.de}

\author{Conglang Zhang}
\authornotemark[1]
\affiliation{%
  \institution{Wuhan University}
  \city{Wuhan}
  \country{China}
}
\affiliation{%
  \institution{Huawei Technologies Ltd.}
  \city{Wuhan}
  \country{China}
}
\email{ConglangZhang@whu.edu.cn}

\author{Chuanheng Fu}
\affiliation{%
  \institution{Wuhan University}
  \city{Wuhan}
  \country{China}
}
\email{fuchuanheng@whu.edu.cn}

\author{Zihao Yang}
\affiliation{%
  \institution{University of Science and Technology of China}
  \city{Hefei}
  \country{China}
}
\email{zihaoy798@mail.ustc.edu.cn}

\author{Kaixuan Zhou}
\authornote{Project lead}
\affiliation{%
  \institution{Huawei Technologies Ltd.}
  \city{Wuhan}
  \country{China}
}
\email{zhoukaixuan2@huawei.com}

\author{Yizhi Zhang}
\affiliation{%
  \institution{Huawei Technologies Ltd.}
  \city{Wuhan}
  \country{China}
}
\email{zhangyizhi4@huawei.com}

\author{Yanfeng Zhang}
\authornote{Corresponding authors.}
\affiliation{%
  \institution{Huawei Hilbert Research Center}
  \city{Munich}
  \country{Germany}
}
\email{zhangyanfeng8@huawei.com}

\author{Mingwei Sun}
\affiliation{%
  \institution{Wuhan University}
  \city{Wuhan}
  \country{China}
}
\email{mingweis@whu.edu.cn}

\author{Zhen Dong}
\affiliation{%
  \institution{Wuhan University}
  \city{Wuhan}
  \country{China}
}
\email{dongzhenwhu@whu.edu.cn}

\author{Xiaoxiao Long}
\affiliation{%
  \institution{Nanjing University}
  \city{Nanjing}
  \country{China}
}
\email{xxlong@nju.edu.cn}

\author{Zengmao Wang}
\authornotemark[3]
\affiliation{%
  \institution{Wuhan University}
  \city{Wuhan}
  \country{China}
}
\email{wangzengmao@whu.edu.cn}

\author{Liqiu Meng}
\affiliation{%
  \institution{Technical University of Munich}
  \city{Munich}
  \country{Germany}
}
\email{liqiu.meng@tum.de}

\renewcommand{\shortauthors}{Xuyang Chen et al.}

\input{sec/0_abstract}



\maketitle

\begin{figure*}[t]
  \centering
  \includegraphics[width=\textwidth]{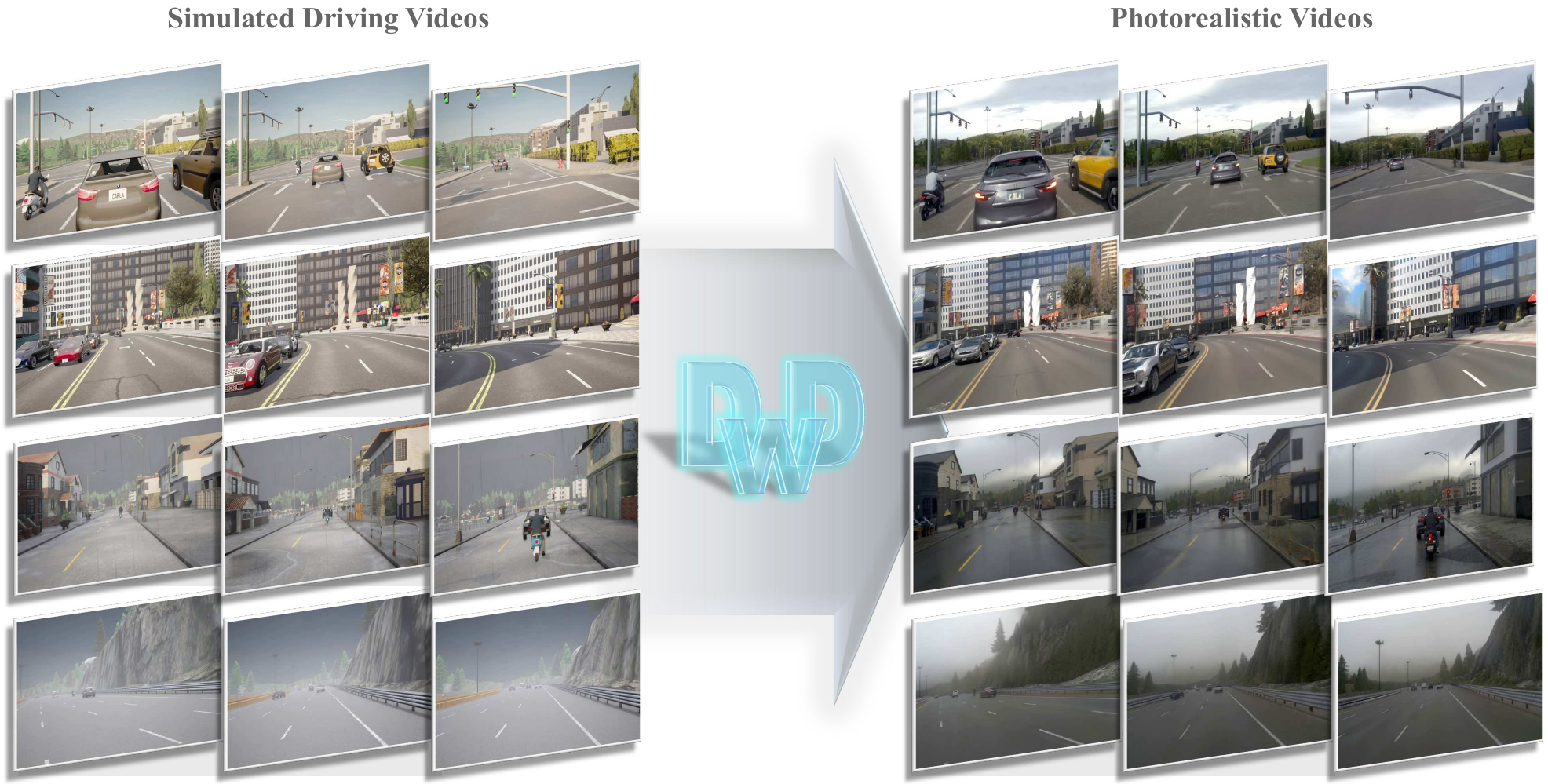}
  \caption{\modelfull{} (\model{}) achieves photorealistic simulation-to-real video translation with superior structural consistency. As shown, our method generalizes remarkably well across diverse scenarios, ranging from town and urban environments to challenging weather conditions such as rain and frost.}
  \Description{Side-by-side examples of four simulated driving sequences and their photorealistic translations. The examples cover urban streets, a town road, rainy conditions, and a frosty highway while preserving the road layout and vehicle motion.}
  \label{fig:teaser}
\end{figure*}

\input{sec/1_intro}
\input{sec/2_related_work}
\input{sec/3_methods}
\input{sec/4_experiments}
\input{sec/5_conclusion}

\input{sec/6_extrafigure}
\bibliographystyle{ACM-Reference-Format}
\bibliography{references}

\input{sec/7_appendix}

\end{document}

%% file: sec/0_abstract.tex
\begin{abstract}
Simulation-to-Real (Sim-to-Real) video translation is essential for bridging the domain gap in autonomous driving, yet existing methods built on controllable video diffusion face a fundamental \textit{Consis\-tency-Realism Dilemma}: low-level conditioning signals (e.g., edges, blurred images) ensure structural control but ``bake in'' synthetic textures, whereas high-level priors (e.g., depth, semantics) enable photorealism but discard the structural detail needed for consistent guidance.
We observe that Vision Foundation Model (VFM) features, specifically DINOv3, inherently encode both low-level structure and high-level semantics within a unified latent space, making them a natural bridge that sidesteps this trade-off entirely.
Building on this insight, we present \modelfull{} (\model{}), a controllable video diffusion framework that conditions generation on DINO features through three targeted designs. Notably, \model{} is trained exclusively on real-world driving videos without requiring any simulated data or sim-real paired samples.
\textit{Minor Components Pruning} applies PCA-based projection with stochastic \textit{Random Channel Tail Drop} to suppress the high-frequency texture details responsible for synthetic artifact leakage, while preserving essential structural cues.
\textit{Spatial Resolution Enhancer} compensates for DINO's $16\times$ spatial downsampling by processing inputs at higher resolution and aligning them to the diffusion backbone via a learnable \textit{Spatial Alignment Module}, recovering fine-grained boundary control.
\textit{Causal Temporal Aggregator} replaces naive keyframe sampling with causal-convolution-based downsampling that preserves historical motion context, mitigating motion blur and enhancing temporal stability.
Experiments show state-of-the-art fidelity (sKID, sFID) and temporal consistency, with competitive photorealism and structural consistency; see our \href{https://albertchen98.github.io/DwD-project/}{project page}.
\end{abstract}

%% file: sec/1_intro.tex
\section{Introduction}

Closed-loop validation is critical for autonomous driving safety, yet applying it in the real world is risky and inefficient due to the difficulty of encountering critical corner cases. Simulation offers a scalable alternative, but creating photorealistic assets remains costly. Consequently, simulators like CARLA~\cite{dosovitskiy2017carla} often compromise on visual fidelity, resulting in a significant domain gap that necessitates robust Sim-to-Real transfer techniques.

\begin{figure}[htbp]
    \centering
    \includegraphics[width=\linewidth]{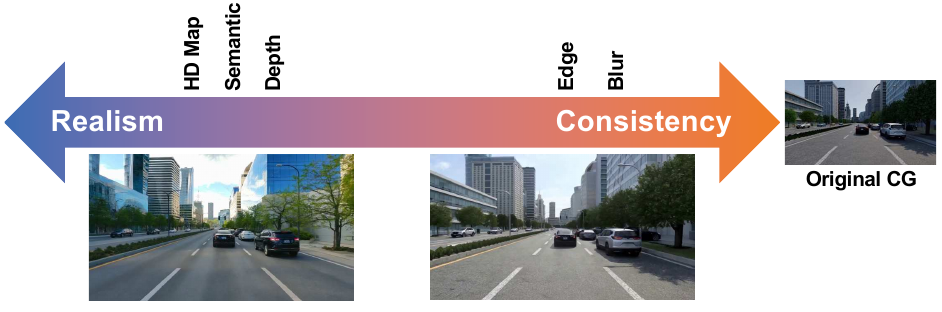}
    \caption{The Consistency-Realism Dilemma. Low-level signals ensure structural control but bake in synthetic textures, while high-level priors enable photorealism but lack fine-grained structural guidance.}
    \Description{A horizontal spectrum from realism to consistency. Depth and segmentation controls lie toward realism, blur and edge controls lie toward consistency, and example driving images illustrate the visual trade-off relative to the original computer-generated input.}
    \label{fig:sim_real_dilemma}
\end{figure}

Among recent generative paradigms, video diffusion models augmented with structural control mechanisms (e.g., ControlNet~\cite{zhang2023adding}) have emerged as a highly promising direction for Sim-to-Real transfer~\cite{zhang2023controlvideo,wang2023videocomposer,guo2024sparsectrl,he2024cameractrl,ma2024trailblazer,xi2025omnivdiff,jiang2025vace}. Within this paradigm, the choice of intermediate representation---the conditioning signal that bridges simulation and reality---becomes the decisive factor for generation quality.

However, existing intermediate representations face a fundamental trade-off that we term the \textbf{Consistency-Realism Dilemma} (Fig.~\ref{fig:sim_real_dilemma}). Current widely adopted representations~\cite{alhaija2025cosmos,ali2025world,gao2023magicdrive,gao2025magicdrive} fail to simultaneously achieve high photorealism and spatial consistency:
\begin{itemize}[leftmargin=*, nosep]
\item \textit{Low-level signals (e.g., edges, blurred images)} provide strong structural guidance but impose rigid constraints that ``bake in'' synthetic textures, hindering photorealistic generation.

\item \textit{High-level signals (e.g., depth, semantics, HDMaps)} enable photorealistic generation but heavily discard structural information, creating ambiguity that frequently leads to generative hallucinations such as spurious or incorrect lane lines.
\end{itemize}

A natural attempt to resolve this dilemma is combining multiple control signals jointly. Yet multi-control frameworks face their own obstacles: jointly training multiple ControlNets incurs prohibitive GPU memory costs and suffers from gradient conflicts between heterogeneous modalities~\cite{chen2023control, hu2023videocontrolnet}, while combining separately trained adapters during inference leads to feature interference and incoherent fusion~\cite{Zhao2023}.

Departing from such multi-branch designs, we propose a fundamentally different conditioning strategy that leverages Vision Foundation Models (VFMs), specifically DINOv3~\cite{simeoni2025dinov3}, as a \textit{unified} bridge between simulation and reality. DINO features inherently encode a rich spectrum of scene properties---from low-level edges and textures to high-level depth and semantics---within a single latent space~\cite{kirillov2023segment,wang2025vggt,wang2025pi,yang2024depth,xu2025pixel,zheng2025diffusion}, naturally fusing structural and semantic information without the feature competition and gradient conflicts of multi-branch architectures.

However, harnessing DINO features for Sim-to-Real transfer requires overcoming three technical challenges. \textbf{First}, DINO representations can act as powerful autoencoders~\cite{zheng2025diffusion,jia2025dino,shi2025latent}, causing \textit{textural information leakage}: the model memorizes and reproduces synthetic textures instead of generating realistic details. \textbf{Second}, DINO features are spatially down-sampled by $16\times$, losing critical structural details along object boundaries. \textbf{Third}, DINO features are extracted frame-wise with significant temporal redundancy; naive keyframe sampling~\cite{burgert2025go,russell2025gaia} disrupts motion continuity, leading to temporal aliasing.

To address these challenges, we propose \textbf{\modelfull{} (\model{})}, a controllable video diffusion framework that leverages DINO features to balance visual realism and control consistency through three targeted designs:
(1)~\textit{Minor Components Pruning} applies PCA-based projection with stochastic \textit{Random Channel Tail Drop} to suppress texture leakage while preserving structural cues;
(2)~\textit{Spatial Resolution Enhancer} compensates for the $16\times$ downsampling by processing inputs at higher resolution and aligning them to the diffusion backbone via a learnable \textit{Spatial Alignment Module};
(3)~\textit{Causal Temporal Aggregator} replaces naive keyframe sampling with causal-convolution-based downsampling that preserves historical motion context, mitigating motion blur and enhancing temporal stability.
Extensive experiments demonstrate that \model{} achieves state-of-the-art fidelity (sKID, sFID) and temporal consistency, while maintaining photorealism and structural consistency competitive with the best specialized baselines.

%% file: sec/2_related_work.tex
\section{Related Work}
\begin{figure*}[!t]
    \centering
    \includegraphics[width=\linewidth]{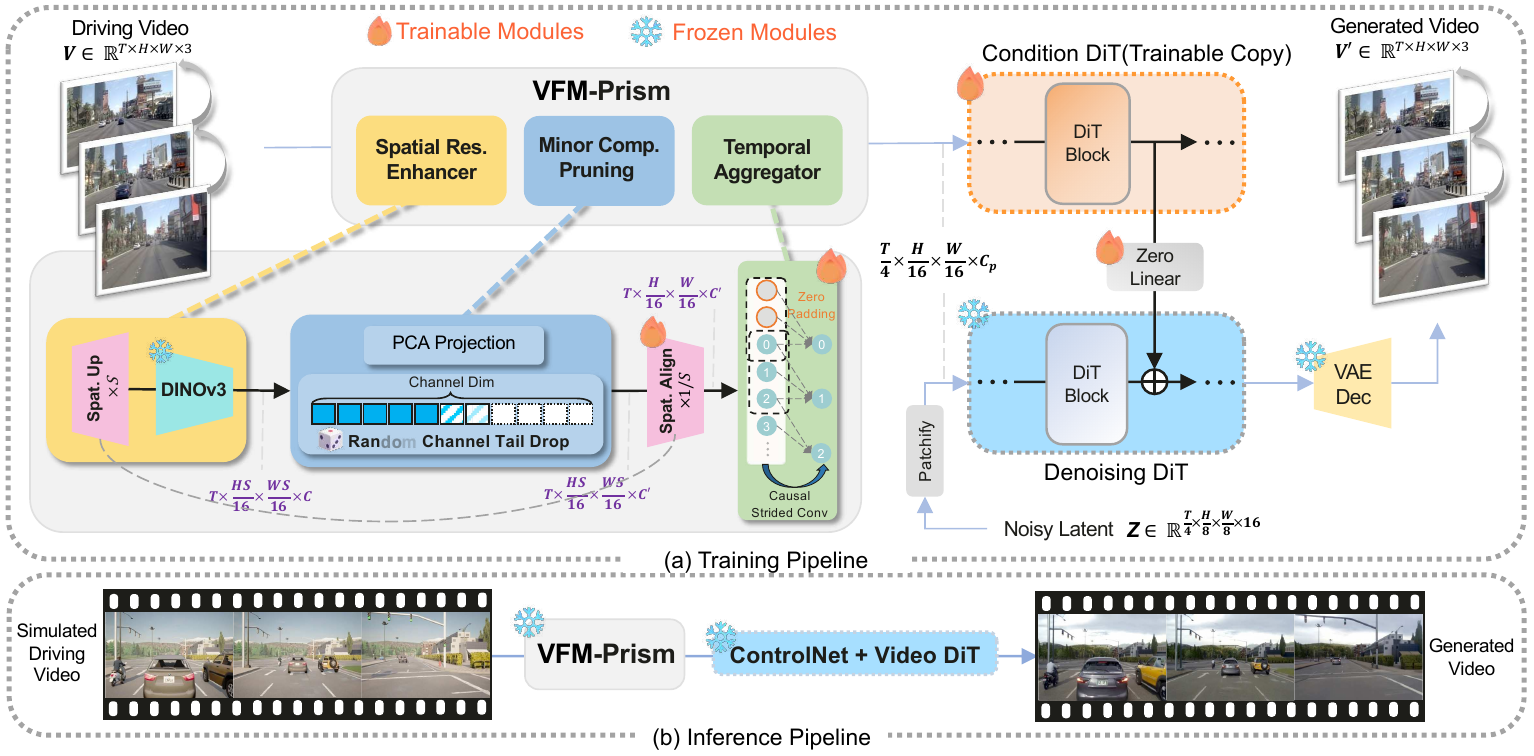}
    \caption{The framework of \model{}. (a) Training: The model is trained on real-world driving videos using a controllable diffusion architecture. The core module, VFM-Prism, processes DINOv3 features through Spatial Resolution Enhancement, Minor Components Pruning, and Causal Temporal Aggregation. These refined features are injected via a Control Branch to guide the reconstruction of the original video. (b) Inference: The model performs Sim-to-Real translation using synthetic inputs. By leveraging the domain-invariant structural features extracted by VFM-Prism (specifically via PCA-based pruning to mitigate texture leakage), \model{} generates high-fidelity photorealistic videos that strictly preserve the simulation's geometric layout.}
    \Description{Two-part architecture diagram. During training, frames from a real driving video pass through DINOv3 and VFM-Prism before conditioning a diffusion transformer that reconstructs the video. During inference, the same feature-processing and conditioning path converts a simulated driving video into a photorealistic video while the diffusion backbone remains frozen.}
    \label{fig:mainfigure}
\end{figure*}
\subsection{Sim-to-Real Transfer}
GAN-based methods such as CycleGAN~\cite{zhu2017unpaired} and CUT~\cite{park2020contrastive} pioneered unpaired Sim-to-Real translation by aligning simulated and real-world distributions through adversarial training. While incorporating simulator-derived G-buffers (e.g., depth, normals) can constrain geometric consistency~\cite{richter2022enhancing}, these approaches frequently suffer from training instability, geometric distortions, and limited texture fidelity.

Diffusion models have largely superseded GANs by offering stable training and superior fine-grained fidelity. SDEdit~\cite{meng2021sdedit} enables transfer via SDE inversion~\cite{song2020denoising}, injecting photorealistic priors while preserving semantic structure. Zero-shot video extensions such as Pix2Video~\cite{ceylan2023pix2video}, TokenFlow~\cite{geyer2023tokenflow}, and FLATTEN~\cite{cong2023flatten} enforce temporal coherence through feature propagation or optical flow injection. Alternative approaches~\cite{zhang2025scaling,liu2025tc} attempt Sim-to-Real transfer via diffusion-based relighting. However, all of these methods rely on implicit guidance (attention manipulation or illumination modification) rather than explicit structural control, and consequently struggle to disentangle semantic structure from synthetic style---often resulting in severe ``texture baking'' or geometric drift.

\subsection{Controllable Video Diffusion for Driving}
ControlNet~\cite{zhang2023adding} introduced explicit spatial conditioning to decouple geometry from appearance in diffusion models. With the evolution of Video Diffusion Transformers (DiTs)~\cite{peebles2023scalable}, controllable video generation has shifted toward architectures that inherently model temporal dynamics~\cite{zhang2023controlvideo,wang2023videocomposer,xi2025omnivdiff,jiang2025vace}, offering temporal consistency superior to zero-shot adaptations.

In the autonomous driving domain, methods such as MagicDrive~\cite{gao2023magicdrive,gao2025magicdrive} and DriveDreamer~\cite{wang2024drivedreamer,zhao2025drivedreamer} utilize sparse high-level priors (e.g., HDMaps, 3D bounding boxes) to govern road topology and traffic placement. WoVoGen~\cite{lu2024wovogen} extends this to volumetric representations via 3D occupancy maps. While effective for layout-level control, these spatially sparse signals leave extensive unconstrained regions prone to generative hallucinations, such as erroneous lane markings or spurious vehicles.

Recent frameworks such as Cosmos Transfer~\cite{alhaija2025cosmos,ali2025world} attempt to unify these paradigms by supporting both low-level (e.g., edges, blurred video) and high-level (e.g., depth, semantics, HDMaps) conditions. However, any single modality inevitably encounters the \textit{Consistency-Realism Dilemma} described in Sec.~1. Furthermore, fusing heterogeneous signals introduces significant challenges: multi-adapter approaches limited to similar modalities~\cite{gao2025longvie} avoid conflicts but sacrifice complementarity, while naive fusion of heterogeneous inputs suffers from prohibitive training costs and instability~\cite{xi2025ctrlvdiff}. These difficulties motivate the search for a unified representation that inherently encapsulates both structural and semantic information within a single control signal.

\subsection{Vision Foundation Models as Control Signals}
Vision Foundation Models (VFMs) such as DINOv2~\cite{oquab2023dinov2} and DINOv3~\cite{simeoni2025dinov3} learn rich visual representations that encode semantics, geometry, and texture without task-specific supervision. Their versatility has been demonstrated across segmentation~\cite{kirillov2023segment}, depth estimation~\cite{yang2024depth,wang2025vggt}, and image reconstruction~\cite{zheng2025diffusion,jia2025dino,shi2025latent}. In the generative domain, VFM features have been adopted for object-level appearance control: AnyDoor~\cite{chen2024anydoor} and DIVE~\cite{jiang2025dive} leverage DINO features to guide identity-preserving object insertion and editing.

However, existing VFM-based control methods target object-level manipulation and have not been extended to scene-level Sim-to-Real transfer. Moreover, naive injection of DINO features leads to texture leakage, where the model reproduces source-domain textures rather than generating realistic details. Our work addresses this gap by introducing a principled feature processing pipeline that transforms DINO features into a unified conditioning signal for scene-level Sim-to-Real video translation.

%% file: sec/3_methods.tex
\section{Method}

\subsection{Overview}\label{subsec:31}
Given a synthetic driving video $\boldsymbol{V}_\text{sim} \in \mathbb{R}^{T \times H \times W \times 3}$, \model{} generates a photorealistic counterpart $\boldsymbol{V}_\text{real}$ that preserves the simulation's geometric layout. The core insight is to leverage DINOv3~\cite{simeoni2025dinov3} features as a unified conditioning signal that inherently encodes both structural and semantic information, eliminating the need for multi-branch control architectures.

As shown in Fig.~\ref{fig:mainfigure}, \model{} builds upon Cosmos-Predict2.5~\cite{ali2025world} (WAN2.1 3D VAE~\cite{wan2025wan} + DiT blocks), augmented with a dedicated control branch comprising conditioning DiT blocks initialized from the denoising DiT weights. The input video is first processed by \textit{VFM-Prism} (Sec.~\ref{subsec:33}), which extracts DINOv3 features and transforms them into conditioning signals through spatial enhancement, PCA-based pruning, and causal temporal aggregation. These refined features are injected into the conditioning DiT blocks, whose outputs are modulated into the denoising DiT at uniform intervals following VACE~\cite{jiang2025vace}.

During training, only the conditioning DiT is updated while the denoising DiT remains frozen. Crucially, \model{} is trained exclusively on real-world driving videos without requiring simulated data or sim-real paired samples: DINOv3 maps both domains into a shared structural representation, so the control branch trained on real data generalizes directly to synthetic inputs at inference (Sec.~\ref{subsec:34}).

\subsection{VFM-Prism: Feature Processing Pipeline} \label{subsec:33}
Three challenges must be addressed before DINO features can serve as effective Sim-to-Real conditioning: (1)~\textit{texture leakage} from raw features encoding synthetic textures; (2)~\textit{spatial information loss} from $16\times$ downsampling; and (3)~\textit{temporal dimension mismatch} from frame-wise processing. We address these through three modules collectively termed \textit{VFM-Prism} (Fig.~\ref{fig:mainfigure}(a)). The full pipeline proceeds as: (1)~input upscaling + DINOv3 encoding to obtain high-resolution features, (2)~PCA-based pruning to suppress texture leakage, and (3)~the Spatial Alignment Module with Causal Temporal Aggregator to match the conditioning DiT's spatial and temporal resolution. Below we describe each module.

\subsubsection{Spatial Resolution Enhancer}
DINOv3's $16\times$ spatial compression discards fine-grained details critical for driving---lane markings, traffic signs, and object boundaries. To recover this information, we upsample the input by factor $S$ prior to DINOv3 encoding, yielding feature maps $\boldsymbol{Z}_c \in \mathbb{R}^{T \times \frac{HS}{16} \times \frac{WS}{16} \times C}$ with $S\times$ finer spatial detail. Unlike previous VFMs~\cite{radford2021learning,oquab2023dinov2,tschannen2025siglip} that degrade at non-native resolutions due to fixed positional embeddings, DINOv3 remains robust under input upscaling thanks to its RoPE-box jittering mechanism.

\begin{figure}[htbp]
    \centering
    \includegraphics[width=\linewidth]{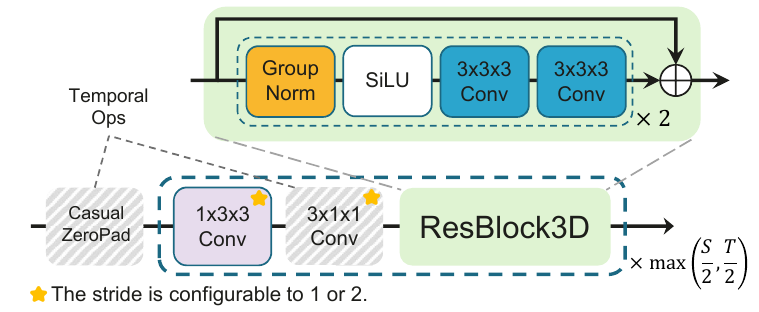}
    \caption{Architecture of the Spatial Alignment Module and Causal Temporal Aggregator.}
    \Description{Block diagram of spatial and temporal feature alignment. Convolution and residual blocks reduce spatial resolution, while a causal convolution aggregates adjacent frames and reduces the temporal dimension before producing the aligned conditioning features.}
    \label{fig:stalign}
\end{figure}

A \textbf{Spatial Alignment Module} (Fig.~\ref{fig:stalign}) then downsamples $\boldsymbol{Z}_c$ to the $\frac{H}{16} \times \frac{W}{16}$ resolution expected by the conditioning DiT, using $\log_2 S$ strided-convolution stages interleaved with residual blocks. We opt for input-space upscaling over post-hoc feature upsampling methods (e.g., FeatUp~\cite{fu2024featup}, AnyUp~\cite{wimmer2025anyup}). As shown in the appendix, these methods incorporate high-resolution RGB pixels as guidance, which introduces false semantic boundaries from pixel-level texture differences and fails to recover fine-grained structural details even at high upsampling factors, leading to unstable training and degraded Sim-to-Real generation quality.

\begin{figure*}[!tb]
    \centering
    \includegraphics[width=\textwidth]{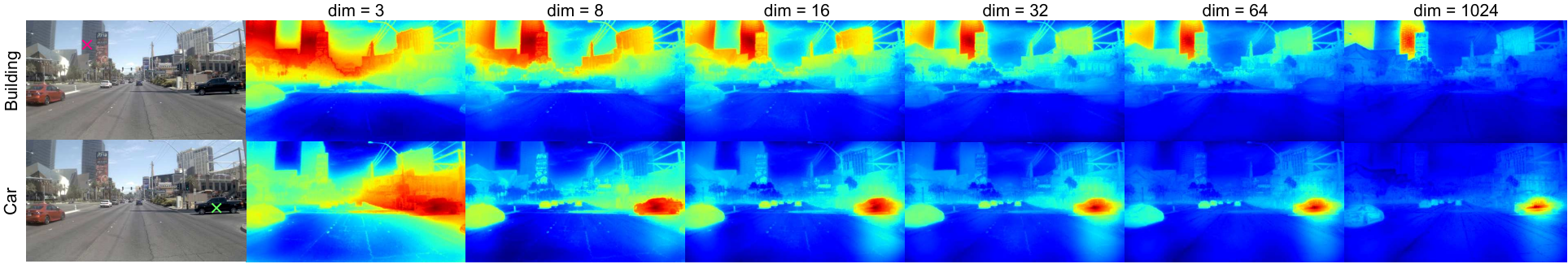}
    \caption{PCA and similarity visualization of DINOv3 features. Lower-dimensional PCA components encode coarse semantic layouts, whereas increasing the dimensions leads to the refinement of high-frequency details. Similarity maps are shown for two anchor points (top: \textcolor{red}{$\times$} on building, bottom: \textcolor{green}{$\times$} on one car).}
    \Description{A driving image followed by PCA visualizations for increasing numbers of retained components and two rows of similarity heat maps. Low-dimensional features preserve broad semantic regions; higher dimensions progressively reveal fine texture. The heat maps highlight pixels similar to an anchor on a building and an anchor on a car.}
    \label{fig:similarity_pca}
\end{figure*}

\subsubsection{Minor Components Pruning}
Raw DINO features entangle structural geometry with domain-specific textures. When used directly as conditioning, this entanglement causes \textit{texture leakage}: the diffusion model memorizes and reproduces CG surface details instead of generating realistic textures, as the conditioning signal effectively serves as an auto-encoder~\cite{zheng2025diffusion,jia2025dino,shi2025latent}.

To disentangle structure from texture, we project DINOv3 features in $\mathbb{R}^C$ onto a $k$-dimensional PCA subspace. Let $\boldsymbol{W} \in \mathbb{R}^{C \times k}$ denote the leading $k$ eigenvectors of the feature covariance matrix, pre-computed over the training set via Incremental PCA. Given a feature vector $\boldsymbol{z} \in \mathbb{R}^C$, the pruned representation is $\boldsymbol{z}' = \boldsymbol{W}^\top (\boldsymbol{z} - \boldsymbol{\mu})$, where $\boldsymbol{\mu}$ is the global mean. As visualized in Fig.~\ref{fig:similarity_pca}, leading PCA components encode coarse semantic layouts, while minor components capture high-frequency textures. The truncation $k$ governs the structure-texture trade-off: very low $k$ (e.g., 3) collapses spatial distinctions, while high $k$ (e.g., 64) retains texture leakage; a moderate range ($k \in \{8, 16, 32\}$) balances both. A further consequence of this pruning is that it closes the sim-real gap in feature space: by stripping domain-specific textures, pruned features from simulated and real inputs become structurally aligned, which is the key enabler of real-only training (Sec.~\ref{subsec:34}).

Since the boundary between structural and textural information in PCA components is not sharply defined, we propose \textit{Random Channel Tail Drop} to make the model robust across truncation levels. During training, we sample an active channel count $k'$ uniformly from a discrete set $\mathcal{K} = \{k_1, \dots, k_m\}$ and apply a binary mask $\mathbf{M} \in \{0, 1\}^{k_m}$ with $\mathbf{M}_{i} = 1$ if $i \le k'$, zeroing out trailing dimensions. This stochastic strategy prevents the model from relying on any specific PCA dimension and instead encourages it to prioritize semantic cues present across all truncation levels. A key practical advantage is that the truncation dimension $k$ becomes a tunable parameter at inference time without retraining: lower $k$ encourages stronger photorealism, while higher $k$ preserves finer structural details.

\subsubsection{Causal Temporal Aggregator}
DINOv3 processes each frame independently, producing $T$ feature maps for a $T$-frame video. However, the conditioning DiT operates on temporally compressed latents (the WAN2.1 3D VAE compresses the temporal dimension by $4\times$), creating a dimension mismatch that must be resolved. Prior approaches resort to heuristics: keyframe sampling~\cite{burgert2025go,hwang2025cross} discards inter-frame information, while bilinear interpolation~\cite{russell2025gaia} introduces temporal smearing. Both disrupt motion continuity and cause visible artifacts under fast camera movement or jittering.

We address this by integrating causal convolution blocks into the Spatial Alignment Module (Fig.~\ref{fig:stalign}), inserting temporal convolution layers with stride 4 after the spatial downsampling stages. Following the WAN2.1 3D VAE convention, we adopt causal masking so that each output latent at time $t$ depends only on frames $\{1, \dots, t\}$, with zero-padding at boundaries. Unlike keyframe sampling---which discards intermediate frames and introduces temporal aliasing under fast motion---or bilinear interpolation---which blurs across frames---our learned aggregator preserves motion dynamics by incorporating all frames within each temporal window into the downsampled representation.

\subsection{Sim-to-Real Inference} \label{subsec:34}
A key design property of \model{} is that the control branch is trained exclusively on real-world driving videos, yet generalizes to synthetic inputs at inference without any sim-real paired data. This is made possible by Minor Components Pruning (Sec.~\ref{subsec:33}): by projecting DINOv3 features onto a low-dimensional PCA subspace, domain-specific textures---the primary source of the sim-real gap---are discarded, leaving only structural and semantic cues that are shared across both domains. Consequently, the conditioning DiT learns to map pruned features to photorealistic outputs regardless of whether those features originate from real or simulated frames.

At inference (Fig.~\ref{fig:mainfigure}(b)), given a synthetic video $\boldsymbol{V}_\text{sim}$, the pipeline proceeds as follows: (1)~each frame is spatially upscaled by factor $S$; (2)~the upscaled frames are encoded by DINOv3 to obtain high-resolution feature maps; (3)~the features are projected onto the pre-computed PCA basis and truncated to $k$ dimensions; (4)~the Spatial Alignment Module downsamples the features to the conditioning DiT's spatial resolution, while the Causal Temporal Aggregator compresses the temporal dimension by $4\times$ via causal convolutions; (5)~the resulting conditioning latents modulate the frozen denoising DiT, which generates $\boldsymbol{V}_\text{real}$ from Gaussian noise through iterative denoising.

Thanks to \textit{Random Channel Tail Drop} during training, the truncation dimension $k$ serves as an inference-time control knob without retraining: lower $k$ suppresses more texture detail and encourages stronger photorealism, while higher $k$ preserves finer structural cues at the cost of potential texture leakage. This flexibility allows practitioners to adapt the realism--consistency trade-off to downstream task requirements (e.g., aggressive realism for perception training data, or conservative structural preservation for safety-critical validation).

%% file: sec/4_experiments.tex
\section{Experiments}
\subsection{Experiment Setup}

\subsubsection{Training Data}
We train \model{} on the nuPlan dataset~\cite{caesar2021nuplan}. An automated curation pipeline using Qwen3-VL~\cite{Qwen3-VL} filters out visually degraded samples and generates captions, yielding 6,000 curated videos of 200 frames each. A fixed global PCA basis is computed via Incremental PCA over one randomly sampled frame per video to resolve eigenvector sign ambiguity.

\subsubsection{Training Configuration}
Videos are resized to $1280 \times 704$ and randomly cropped into 93-frame chunks. Only the conditioning DiT is updated (denoising DiT frozen). Training runs for 20k iterations (batch size 1, lr $5 \times 10^{-5}$) on 8 GPUs (48GB each) with Context Parallelism.

\subsubsection{Evaluation Data}
We synthesize 149 videos (avg.\ 200 frames) using CARLA~\cite{dosovitskiy2017carla} across various maps with randomized routes and dynamic agents. Additional custom-built scenes evaluate generalizability.

\subsubsection{Baselines}
We compare against TC-Light~\cite{liu2025tc} (training-free relighting), FRESCO~\cite{yang2024fresco} (zero-shot style transfer with nuPlan reference images), and Cosmos-Transfer2.5~\cite{ali2025world} with various modalities (Edge, Blur, Depth, Seg, and multi-control). All video generation baselines are fine-tuned on the same training set for fair comparison.

\begin{figure*}[htbp]
    \centering
    \includegraphics[width=\linewidth]{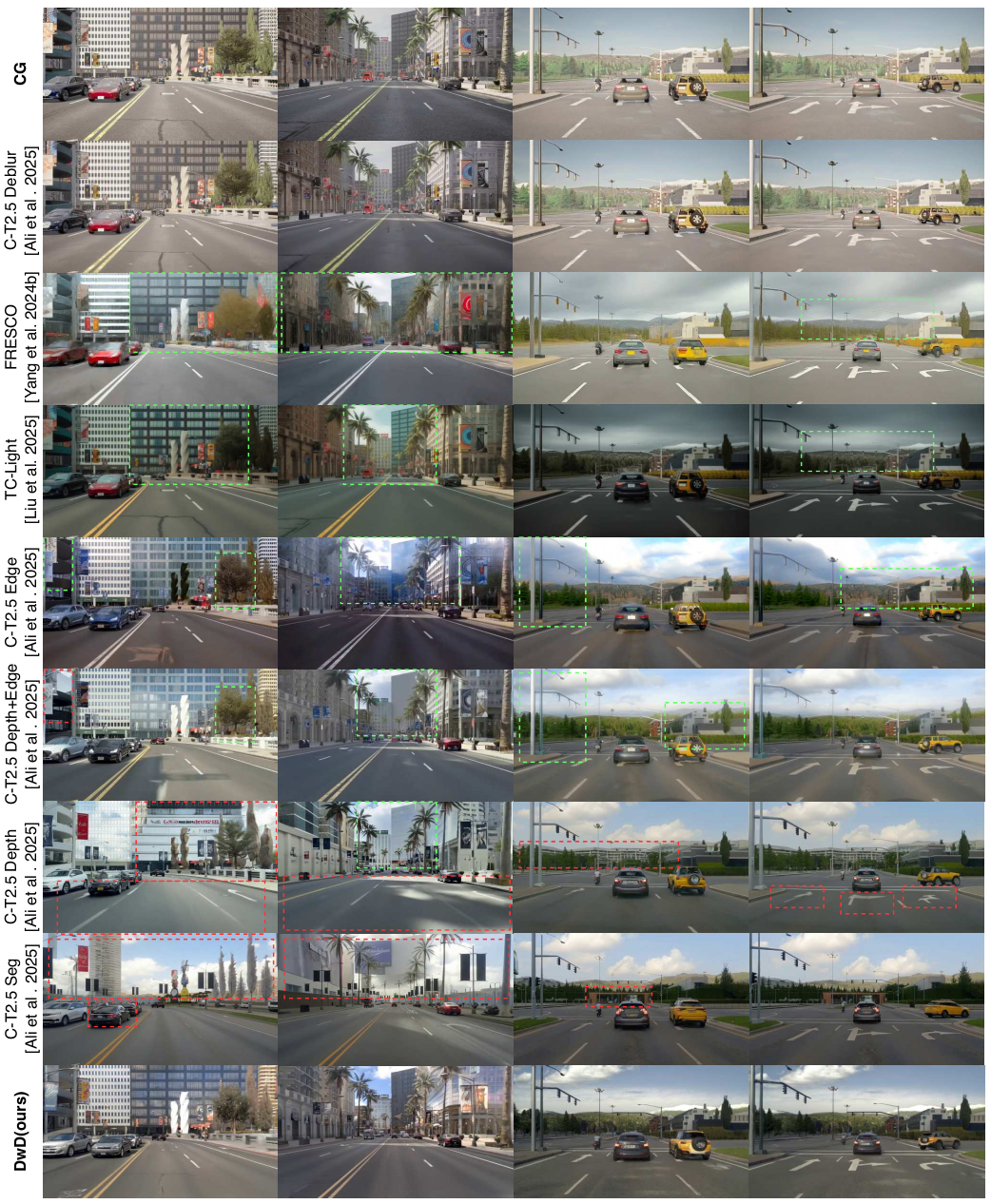}
    \caption{Qualitative comparison with state-of-the-art methods. \textcolor{red}{Red boxes}: \textcolor{red}{inconsistencies} w.r.t.\ the CG input. \textcolor{green}{Green boxes}: \textcolor{green}{low-fidelity textures}. Best viewed zoomed in.}
    \Description{A grid comparing the computer-generated input, several transfer baselines, and the proposed method across urban driving scenes. Red boxes mark changed lane geometry or other structural errors, while green boxes mark synthetic-looking texture. The proposed method preserves the input structure while producing realistic road and building textures.}
    \label{fig:Qualitative_Comparison}
\end{figure*}

\subsection{Evaluation Metrics}
We assess results along four dimensions:
\begin{itemize}[leftmargin=*, nosep]
\item \textit{Visual Fidelity} via \textbf{sFID} and \textbf{sKID} using semantics-aware patch sampling~\cite{richter2022enhancing} (400K matched pairs).
\item \textit{Perceptual Realism} via \textbf{CLIP-Real}~\cite{zeng2025neuralremaster}, defined as $(x^\top t_p) / (x^\top t_n)$ for CLIP embeddings of the output and the positive (``Photo'') and negative (``Game'') prompts.
\item \textit{Temporal Consistency} via \textbf{Motion-S}~\cite{huang2024vbench} and \textbf{WarpSSIM} (SSIM between frames warped by CG-derived optical flow).
\item \textit{Sim2Real Consistency} via \textbf{mIoU} between CG ground truth and segmentation from a Cityscapes-pretrained InternImage~\cite{wang2023internimage}.
\end{itemize}

\subsection{Quantitative Evaluation}
Tab.~\ref{tab:Quantitative_results} summarizes results across all four evaluation dimensions. \model{} is the only method that simultaneously achieves leading scores in fidelity, realism, temporal consistency, and sim-to-real consistency, validating that a single DINOv3-based conditioning signal can resolve the Consistency-Realism Dilemma without multi-branch fusion.

\subsubsection{Fidelity \& Realism}
\model{} achieves sKID of 9.30 and sFID of 21.62---reducing sKID by 56\% and sFID by 34\% over the next best method (C-T2.5 Edge$^\dagger$, sKID 21.22 / sFID 32.59). Neither low-level nor high-level baselines match this balance, mirroring the Consistency-Realism Dilemma in Sec.~1. Low-level conditions (Edge, Blur) retain CG textures that inflate sKID/sFID despite strong structural control. Counter-intuitively, high-level modalities (Depth, Seg) also underperform on fidelity: their lack of structural guidance introduces layout discrepancies penalized by the spatially aligned evaluation protocol. This confirms that fidelity requires \emph{both} realistic textures and precise spatial alignment. For CLIP-Real, \model{} (119.11) matches the best high-level method (C-T2.5 Seg: 119.73), demonstrating equivalent realism without sacrificing structural control, while low-level conditions score near the CG baseline (98.79).

\subsubsection{Temporal Consistency}
\model{} achieves the highest Motion-S (98.94\%) and competitive WarpSSIM (93.07). Methods with higher WarpSSIM---TC-Light (96.59) and C-T2.5 Edge$^\dagger$ (94.37)---achieve this by preserving CG textures more rigidly, which inflates pixel-level frame-to-frame similarity at the expense of realism (their CLIP-Real scores of 102.46 and 112.58 are substantially lower than ours at 119.11). The multi-control variant C-T2.5 Edge+Depth suffers a significant Motion-S drop to 93.97\%, suggesting that heterogeneous signal fusion destabilizes temporal coherence---a failure mode \model{} avoids by using a single conditioning modality.

\subsubsection{Sim-to-Real Consistency}
For mIoU (Tab.~\ref{tab:miou_comparison}), \model{} achieves 0.4967, comparable to Edge (0.5015) and Blur (0.5105), and substantially higher than Seg (0.4132). The slight mIoU advantage of Edge and Blur is attributable to their texture-preserving nature: by retaining CG surface details almost verbatim, their outputs present segmentation boundaries that trivially align with CG ground truth. In contrast, \model{} generates photorealistic textures whose semantic boundaries approximate real-world distributions---for instance, gradual road-to-sidewalk transitions rather than sharp CG edges. A Cityscapes-pretrained segmentor may parse these realistic boundaries slightly differently from rigid CG ground truth, explaining the marginal mIoU gap despite superior visual quality (see per-category analysis in the appendix, where \model{} leads in fine-grained categories such as Fence and Bicycle). Notably, Seg achieves the lowest mIoU (0.4132) despite using semantic segmentation as its conditioning signal, confirming that mIoU in this protocol reflects structural consistency of the generated output rather than conditioning signal quality. The low-resolution variant (pca8 $\times$1: 0.4675) confirms that the Spatial Resolution Enhancer contributes meaningfully to structural preservation.

\begin{table}[htbp]
    \centering
     \caption{Quantitative evaluations. Top results: \colorbox{agc}{Gold}, \colorbox{asc}{Silver}, \colorbox{abc}{Bronze} (also for Tabs.~\ref{tab:ablation},~\ref{tab:miou_comparison}). $\dagger$: fine-tuned on same data as \model{}. C-T2.5: Cosmos-Transfer 2.5.}
    \resizebox{\columnwidth}{!}{
        \begin{tabular}{lccccc}
            \toprule
            Method & Mot-S$\uparrow$ & W-SSIM$\uparrow$ & CLIP-R$\uparrow$ & sKID$\downarrow$ & sFID$\downarrow$ \\
            & ($\times 10^2$) & & ($\times 10^2$) & ($\times 10^3$) & \\
            \midrule
            CG & - & - & 98.79 & 29.59 & 43.65 \\
            \midrule
            FRESCO & 98.37 & 93.32 & 109.92 & 32.55 & 56.28 \\
            \midrule
            TC-Light & \third{98.79} & \first{96.59} & 102.46 & \third{22.12} & 44.47 \\
            \midrule
            C-T2.5 Edge & 98.61 & 93.60 & 112.38 & 22.82 & 37.44 \\
            C-T2.5 Edge$^\dagger$ & 98.73 & \second{94.37} & 112.58 & \second{21.22} & \second{32.59} \\
            C-T2.5 Blur & 98.02 & 88.14 & 100.63 & 29.09 & 39.57 \\
            C-T2.5 Depth & 98.47 & 92.48 & \third{116.50} & 26.98 & 38.44 \\
            C-T2.5 Edge+Depth & 93.97 & \third{93.97} & 112.15 & 23.71 & 36.69 \\
            C-T2.5 Edge+Seg & 98.62 & \third{93.97} & 112.13 & 23.32 & \third{36.62} \\
            C-T2.5 Seg & \second{98.85} & 90.92 & \first{119.73} & 23.81 & 37.20 \\
            \midrule
            \textbf{Ours} & \first{98.94} & 93.07 & \second{119.11}& \first{9.30} & \first{21.62} \\
            \bottomrule
        \end{tabular}
    }
    \label{tab:Quantitative_results}
\end{table}

\begin{table}[htbp]
\centering
\caption{mIoU comparison (avg over 14 categories; per-category in appendix). Edge, Blur, Depth, and Seg denote Cosmos-Transfer 2.5 with the respective control signal. pca8 ($\times$1): low-res ablation variant of \model{}.}
\resizebox{\columnwidth}{!}{
\begin{tabular}{lcccccc}
\toprule
 & pca8 ($\times$1) & Edge & Blur & Depth & Seg & \textbf{Ours} \\
\midrule
\textbf{mIoU} & 0.4675 & \second{0.5015} & \first{0.5105} & 0.4843 & 0.4132 & \third{0.4967} \\
\bottomrule
\end{tabular}
}
\label{tab:miou_comparison}
\end{table}

\subsection{Qualitative Evaluation}

Fig.~\ref{fig:Qualitative_Comparison} visualizes the Consistency-Realism Dilemma across methods. Road markings serve as the most diagnostic indicator: they require both precise structural alignment (to avoid hallucination) and photorealistic texture (to avoid CG artifacts).

High-level conditions (Depth, Seg) hallucinate in structurally ambiguous regions. As highlighted by the red boxes in Fig.~\ref{fig:Qualitative_Comparison}, C-T2.5 Depth generates spurious lane markings that do not exist in the CG input, while C-T2.5 Seg misplaces road boundaries and introduces phantom structures. These errors arise because high-level signals discard fine-grained geometric cues, leaving the diffusion model to ``guess'' structural details from learned priors---priors that frequently conflict with the intended simulation layout.

Low-level conditions (Blur, Edge) exhibit the opposite failure mode. As shown in the green boxes, they faithfully preserve CG textures: flat shading, uniform surface colors, and artificial road patterns are carried over into the output, producing results that remain visually synthetic. C-T2.5 Edge$^\dagger$ improves over the pre-trained Edge variant after fine-tuning on nuPlan data, but the fundamental constraint remains---edge maps impose pixel-level rigidity that prevents the model from generating realistic material properties (e.g., specular reflections, asphalt grain, weathered surfaces).

The multi-control variant C-T2.5 Edge+Depth attempts to combine the strengths of both paradigms. While it improves perceptual realism over C-T2.5 Edge alone, it still retains noticeable CG artifacts and fails to fully reconcile these failure modes. This result aligns with the quantitative finding that multi-control fusion destabilizes generation quality (Motion-S drops to 93.97\%), suggesting that heterogeneous conditioning signals compete rather than complement during the denoising process.

In contrast, \model{} achieves both photorealism and structural fidelity through a single unified conditioning signal. Road markings are rendered with realistic texture while maintaining correct geometry; building facades exhibit plausible material variation; and vegetation appears naturalistic without structural drift. These results confirm that DINOv3 features, processed through VFM-Prism, provide an effective middle ground between low-level rigidity and high-level ambiguity. Additional qualitative comparisons across diverse scenes are provided in the supplementary material.

\subsection{Ablation Studies}
We isolate each VFM-Prism component in Tab.~\ref{tab:ablation} to verify that every design choice contributes to the final performance, building from the simplest configuration ($k{=}8$, $\times 1$) to the full model.

\subsubsection{Naive DINO Injection}
Directly using non-pruned DINO latents yields the worst performance across all configurations in Tab.~\ref{tab:ablation}, with CLIP-R and sFID degrading to levels comparable to conventional conditioning baselines. Qualitatively, the output exhibits severe texture leakage where CG surface details are faithfully reproduced (see appendix), confirming the necessity of Minor Components Pruning.

\subsubsection{Minor Components Pruning ($k$)}
Fidelity improves as $k$ decreases from 32 to 8, since fewer retained components means less texture leakage. However, excessive pruning ($k{=}3$) degrades fidelity as the signal becomes too sparse for complex geometries. We select $k{=}8$ as the optimal trade-off; qualitative visualizations of texture leakage at various $k$ are provided in the appendix. Comparing $k{=}8$ with and without Random Channel Tail Drop (``fixed'' row in Tab.~\ref{tab:ablation}), the stochastic training strategy improves all metrics---most notably CLIP-R (117.95$\to$118.78) and sKID (9.83$\to$9.08)---by encouraging the model to prioritize semantic cues over any specific PCA dimension.

\subsubsection{Spatial Upscaling ($S$)}
$S{=}4$ improves W-SSIM (91.34$\to$92.69) with a slight sKID increase (9.08$\to$9.68). Higher $S$ provides fine-grained geometric cues---e.g., $S{=}1$ mistranslates street lamps into traffic lights (see appendix for visual comparison). The fidelity cost stems from preserving pristine CG edges that diverge from real-video motion blur.

\subsubsection{Causal Temporal Aggregator}
Adding the aggregator to $\times$4 simultaneously improves W-SSIM (92.69$\to$93.07) and fidelity (sKID: 9.68$\to$9.30). Under fast camera jittering, naive keyframe sampling causes motion blur, whereas our aggregator preserves coherence (see appendix). Tab.~\ref{tab:miou_comparison} further confirms that both spatial and temporal components boost structural consistency.

\subsubsection{VFM Backbone: DINOv3 vs.\ DINOv2}
We ablate the choice of VFM backbone by replacing DINOv3 with DINOv2~\cite{oquab2023dinov2} under the keyframe configuration ($k{=}8$, $\times 1$). As shown in Tab.~\ref{tab:ablation}, DINOv2 degrades all metrics, with the largest gaps in visual fidelity: sKID increases from 9.08 to 22.35 and sFID from 21.37 to 34.66. CLIP-R and W-SSIM also decline (118.78$\to$114.88, 91.34$\to$89.02). We attribute the superiority to two factors: (1)~DINOv3 features are sharper, contain significantly less noise, and exhibit superior semantic coherence compared to DINOv2 (cf.\ Fig.~13 in~\cite{simeoni2025dinov3}), providing cleaner inputs for Minor Components Pruning; (2)~the RoPE-box jittering mechanism in DINOv3 enables stable feature extraction under high-resolution inputs, whereas DINOv2---trained with fixed positional embeddings---degrades when the input resolution deviates from its pre-training setting. Note that to align with the VAE latent spatial resolution, the DINOv2's input resolution is $1120\times 616$, far exceeding its native $518\times 518$ pre-training resolution, which further exacerbates its positional encoding mismatch.

\begin{table}[htbp]
\centering
\caption{Ablation studies. $k$: active PCA channels. $\times S$: upscaling factor. +Temp: Causal Temporal Aggregator. Motion-S is omitted due to computational constraints.}
\resizebox{\columnwidth}{!}{%
\begin{tabular}{lcccc}
\toprule
Configuration  & W-SSIM$\uparrow$ & CLIP-R$\uparrow$ & sKID$\downarrow$ & sFID$\downarrow$ \\
 & & ($\times 10^2$) & ($\times 10^3$) & \\
\midrule
Naive (no pruning), $\times 1$    & 90.37 & 103.12 & 21.84 & 43.91 \\
$k{=}3$, $\times 1$               & 90.94 & \colorbox{agc}{119.63} & 10.70 & 23.32 \\
$k{=}8$, $\times 1$               & 91.34 & 118.78 & \colorbox{agc}{9.08} & \colorbox{agc}{21.37} \\
$k{=}8$ (fixed), $\times 1$       & 90.87 & 117.95 & 9.83 & 22.14 \\
$k{=}16$, $\times 1$              & \colorbox{abc}{91.49}& 116.00 & 9.92 & 22.60 \\
$k{=}32$, $\times 1$              & 90.29 & 114.38 & 9.91 & 23.01 \\
\midrule
$k{=}8$, $\times 2$    & 90.73 & 118.38 & 9.79 & 23.06 \\
$k{=}8$, $\times 4$    & \colorbox{asc}{92.69} & \colorbox{abc}{118.89} & \colorbox{abc}{9.68} & \colorbox{abc}{21.92} \\
\midrule
$k{=}8$, $\times 4$, +Temp (final) & \colorbox{agc}{93.07} & \colorbox{asc}{119.11} & \colorbox{asc}{9.30} & \colorbox{asc}{21.62} \\
\midrule
DINOv2, $k{=}8$, $\times 1$      & 89.02 & 114.88 & 22.35 & 34.66 \\
\bottomrule
\end{tabular}%
}
\label{tab:ablation}
\end{table}

\subsection{Robustness in Long-Horizon Generation}\label{subsec:long_horizon}
A practical requirement for simulation-based validation is the ability to generate long driving sequences via autoregressive chunking. We observe that most Cosmos Transfer 2.5 control branches suffer from severe error accumulation after approximately three 93-frame generation cycles, manifesting as color oversaturation and progressive quality degradation. In contrast, \model{} maintains visual fidelity across extended horizons (see appendix and supplementary videos). We attribute this robustness to VFM-Prism conditioning: because DINOv3 features provide a semantically stable anchor at each chunk boundary, the accumulated drift in pixel space is corrected by the conditioning signal at every generation step, preventing the compounding artifacts observed in conventional conditioning approaches.

%% file: sec/5_conclusion.tex
\section{Conclusion and Limitations}
This paper addresses the Consistency-Realism Dilemma in Sim-to-Real video translation by proposing \modelfull{} (\model{}). The key idea is to leverage DINOv3 features---processed through VFM-Prism (Minor Components Pruning, Spatial Resolution Enhancement, and Causal Temporal Aggregation)---as a unified conditioning signal that inherently encodes both structural and semantic information. This eliminates the need for multi-branch control architectures and, crucially, requires only real-world driving videos for training without any sim-real paired data. Experiments on CARLA-to-nuPlan transfer demonstrate that \model{} reduces sKID by 56\% and sFID by 34\% over the strongest baseline while achieving the highest temporal consistency and competitive photorealism.

PCA-based pruning can remove high-frequency details, degrading text on signs and billboards. Moreover, our model uses only 20K training iterations at batch size 1, versus 100K iterations at batch size 64 for each Cosmos Transfer 2.5 branch---a $320\times$ compute gap. Yet \model{} outperforms all of these variants on fidelity, indicating substantial room to improve with scale. Future work will integrate \model{} into closed-loop autonomous-driving pipelines for perception and planning.

%% file: sec/7_appendix.tex
\appendix

\section{Supplementary Video Results}

We provide side-by-side video comparisons of all methods on our \href{https://albertchen98.github.io/DwD-project/}{project page}. We strongly encourage readers to watch these videos to evaluate temporal consistency, motion coherence, and visual quality, as these properties are difficult to assess from static frames alone.

\section{Upsampling using AnyUp}

Post-hoc feature upsampling methods such as AnyUp~\cite{wimmer2025anyup} offer a computationally efficient alternative for obtaining high-resolution DINO features; however, when we train DwD using DINOv3 feature maps upsampled by AnyUp, we observe degraded controllability and visual quality compared with naive input-space upscaling. The root cause is that AnyUp incorporates high-resolution RGB images as guidance, inadvertently re-introducing the high-frequency pixel information that Minor Components Pruning is designed to suppress.

\begin{figure*}[!t]
  \centering
  \includegraphics[width=\textwidth]{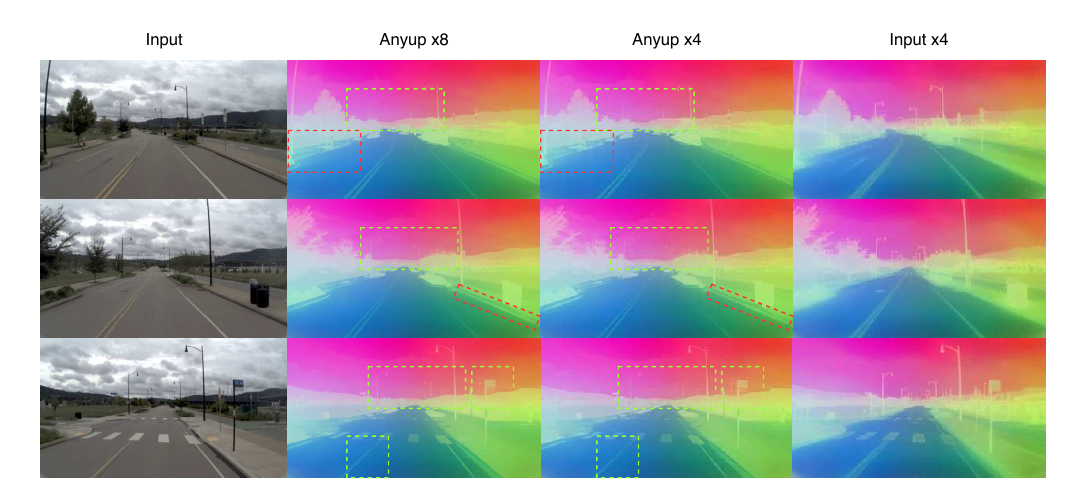}
  \caption{PCA visualization comparing feature maps upsampled via AnyUp ($\times 4$ and $\times 8$) versus naive input upscaling ($\times 4$). The \textcolor{red}{red boxes} highlight semantic inconsistencies introduced by AnyUp, where high-frequency pixel information leads to false semantic boundaries (e.g., misclassifying the curb). The \textcolor{green}{green boxes} demonstrate the failure of AnyUp to recover fine-grained structural details, such as the boundaries of distant street lamps, despite the high upsampling factor.}
  \label{fig:anyup_vis_teaser}
\end{figure*}

Fig.~\ref{fig:anyup_vis_teaser} illustrates two failure modes. First, the \textcolor{red}{red boxes} show that pixel-level guidance causes AnyUp to misinterpret texture differences (e.g., between the curb and neighboring ground) as distinct semantic categories, whereas naive input-space upscaling preserves coherent semantic representations. Second, the \textcolor{green}{green boxes} reveal that AnyUp fails to recover fine-grained structural details even at $8\times$ upsampling---boundaries between distant street lamps and the sky remain indistinct. These observations confirm that post-hoc feature upsampling counteracts the texture suppression of Minor Components Pruning, justifying our choice of input-space upscaling.

\section{Feature-Space Analysis: PCA Pruning Closes the Sim--Real Gap}\label{sec:feature_gap}

\begin{figure}[t]
\centering
\includegraphics[width=0.39\linewidth]{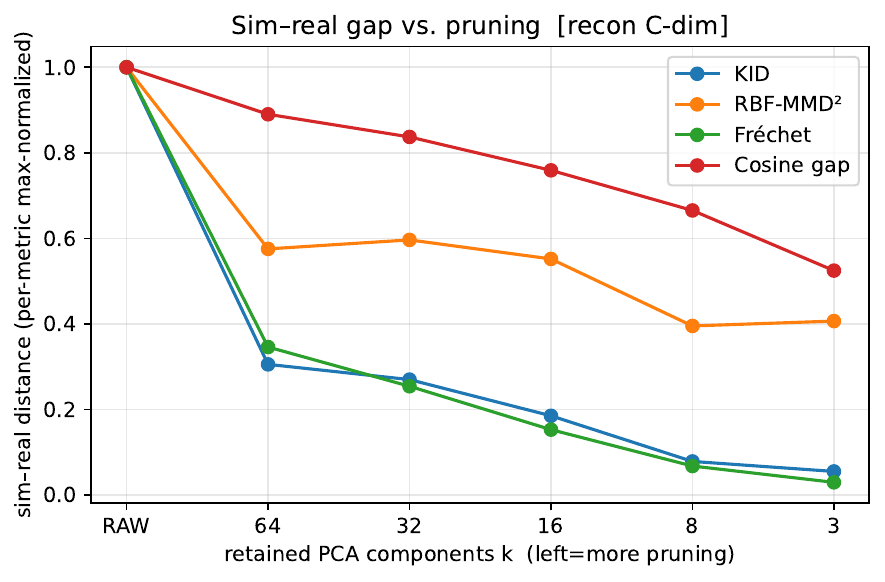}\hfill
\includegraphics[width=0.59\linewidth]{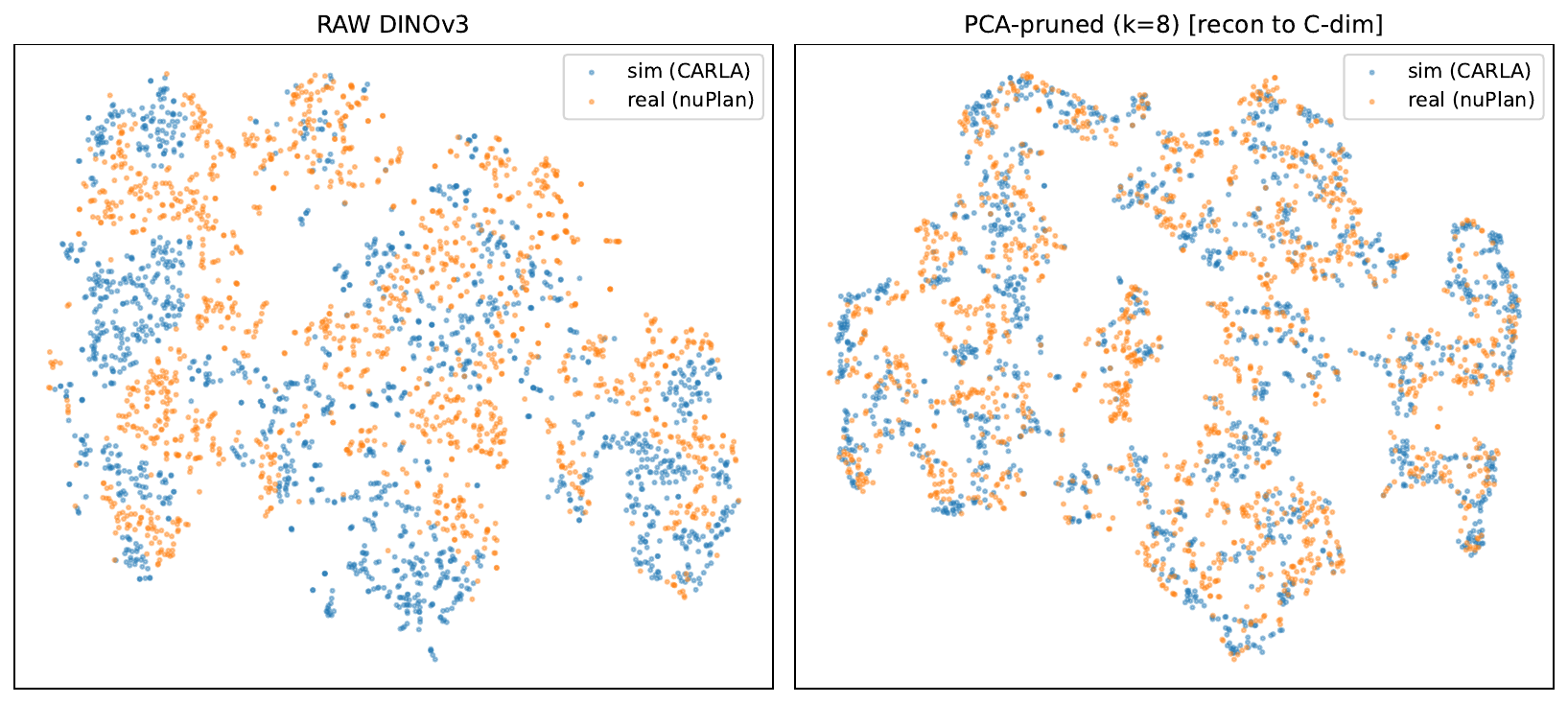}
\caption{PCA pruning closes the sim--real gap on DINOv3 features.
Left: the sim--real distance drops strongly as the retained dimension $k$ shrinks
(KID/Fr\'echet/cosine monotonically; RBF-MMD$^2$ overall). Right: t-SNE of
CARLA (sim) vs.\ nuPlan (real) features---separated on raw features, interleaved
after pruning.}
\Description{Line plot and t-SNE showing that PCA pruning reduces CARLA versus nuPlan
feature-domain separation.}
\label{fig:feature_gap}
\end{figure}

To directly verify the mechanism behind Minor Components Pruning, we measure the domain distance between DINOv3 features extracted from CARLA (sim) and nuPlan (real) frames, before and after pruning (Fig.~\ref{fig:feature_gap}). Pruning from RAW to $k{=}8$ reduces \emph{every} domain-distance metric: KID~\cite{binkowski2018kid} drops from $0.0213$ to $0.0017$ ($-92\%$), RBF-MMD$^2$ from $0.0488$ to $0.0193$ ($-60\%$), Fr\'echet distance from $19.45$ to $1.32$ ($-93\%$), and the cosine gap from $0.80$ to $0.54$. The t-SNE embedding correspondingly changes from clearly separated to interleaved. KID, Fr\'echet, and cosine distances decrease monotonically from RAW to $k{=}8$ (RBF-MMD$^2$ follows the same overall trend), while the main-paper ablation shows that $k{=}3$ is too sparse for complex geometries---together explaining the choice of $k{=}8$.

Importantly, pruning removes \emph{domain-specific texture} rather than structure: structural metrics remain high after pruning (mIoU $0.4967$, close to Edge at $0.5015$; W-SSIM $93.07$). This confirms that the pruned DINOv3 representation is precisely the unified, domain-aligned conditioning signal that enables training exclusively on real-world videos while generalizing to simulated inputs at inference.

\section{Per-Category mIoU Breakdown}

\begin{table}[!t]
\centering
\caption{Per-category mIoU comparison. Edge, Blur, Depth, and Seg denote Cosmos-Transfer 2.5 with the respective control signal. pca8 ($\times$1) is the low-resolution ablation variant of \model{}.}
\resizebox{\columnwidth}{!}{
\begin{tabular}{lcccccc}
\toprule
Category & pca8 ($\times$1) & Edge & Blur & Depth & Seg & \textbf{Ours} \\
\midrule
Road & 0.9574 & \third{0.9591}& \first{0.9636} & 0.9475 & 0.9345 & \second{0.9605} \\
Sidewalk & 0.5464 & \second{0.5914} & \first{0.6332} & 0.5253 & 0.5010 & \third{0.5610} \\
Building & 0.7802 & \first{0.8193} & \second{0.8127} & 0.7725 & 0.5781 & \third{0.8087} \\
Fence & \second{0.0978} & 0.0767 & 0.0790 & 0.0789 & \third{0.0848} & \first{0.1187} \\
Pole & 0.2757 & \first{0.3665} & 0.3214 & \third{0.3548} & 0.2555 & \second{0.3435} \\
Traffic Light & 0.4354 & \first{0.6655} & \second{0.5501} & \third{0.5392} & 0.3174 & 0.4503 \\
Traffic Sign & 0.2014 & \first{0.3838} & \second{0.2862} & 0.2601 & 0.1505 & \third{0.2681} \\
Vegetation & 0.6907 & 0.6833 & \first{0.7210} & \third{0.6936}& 0.5773 & \second{0.7150} \\
Terrain & 0.3171 & \second{0.3561} & \first{0.4476} & \third{0.3284} & 0.3210 & 0.3245 \\
Sky & 0.8945 & 0.8912 & \third{0.9028}& \first{0.9115} & 0.7691 & \second{0.9111} \\
Person & 0.5380 & \first{0.6795} & \second{0.6729} & \third{0.6438} & 0.5828 & 0.6256 \\
Car & 0.7522 & \first{0.7973} & \second{0.7819} & \third{0.7828} & 0.7044 & 0.7756 \\
Bus & \first{0.0989} & 0.0223 & \second{0.0620} & 0.0259& 0.0049 & \third{0.0296}\\
Bicycle & \third{0.0599} & 0.0242 & 0.0132 & 0.0294 & \first{0.0612} & \second{0.0666} \\ \midrule
\textbf{mIoU (Avg)} & 0.4675 & \second{0.5015} & \first{0.5105} & 0.4843 & 0.4132 & \third{0.4967} \\ \bottomrule
\end{tabular}
}\label{tab:miou_full}
\end{table}

Tab.~\ref{tab:miou_full} reports per-category mIoU scores. \model{} ranks first or second in 5 out of 14 categories (Fence, Vegetation, Sky, Bicycle, and Sidewalk) while maintaining balanced performance across the board, achieving an overall mIoU of 0.4967.

The marginal overall gap to the top-performing low-level baselines Edge (0.5015) and Blur (0.5105) requires careful interpretation. Low-level methods preserve CG textures almost verbatim, so the Cityscapes-pretrained segmentor~\cite{wang2023internimage} encounters nearly identical boundary patterns as in the CG ground truth, artificially inflating their mIoU. Their substantially worse sKID and sFID (Tab.~1 in the main paper) corroborate this: high mIoU comes at the expense of visual realism.

In contrast, \model{} generates photorealistic textures whose semantic boundaries approximate real-world distributions---e.g., gradual road-to-sidewalk transitions rather than sharp CG edges---leading to marginal mIoU reductions in large-area categories (Road: 0.9605 vs.\ 0.9636, Sidewalk: 0.5610 vs.\ 0.6332). Crucially, for fine-grained categories such as Fence (0.1187, rank 1st) and Bicycle (0.0666, rank 2nd), \model{} outperforms all baselines: these slender structures are easily lost in edge maps or smoothed out by blur, whereas our DINO-based conditioning retains rich semantic information regardless of spatial extent.

The high-level baseline Seg achieves the lowest overall mIoU (0.4132) despite using semantic segmentation as its conditioning signal, confirming that mIoU in this protocol reflects structural consistency of the \emph{generated output}, not conditioning signal quality.

\section{Downstream Perception Evaluation}\label{sec:downstream}

To assess the practical value of Sim-to-Real translation for perception, we run a \emph{source-only} synthetic$\to$real transfer experiment, the standard protocol for measuring the sim-to-real gap in urban-scene segmentation~\cite{richter2016gta5,ros2016synthia,cordts2016cityscapes}. For each conditioning strategy we train an identical, freshly initialized SegFormer-B2~\cite{xie2021segformer} (ImageNet-pretrained MiT-b2 encoder, randomly initialized head) on that method's translated CARLA frames paired with the \emph{shared} CARLA semantic ground truth, and evaluate on the real Cityscapes validation set~\cite{cordts2016cityscapes}. The only variable across rows is image appearance, which isolates translation quality. We report 19-class mIoU, averaged over 2 seeds.

\begin{table}[t]
\centering\small
\setlength{\tabcolsep}{4pt}
\caption{Source-only sim$\to$real transfer: an identical SegFormer-B2 trained on each condition's translated CARLA frames (shared CARLA GT), evaluated on real Cityscapes val (19-class mIoU, mean of 2 seeds). Only appearance varies across rows. $\dagger$: fine-tuned on the same nuPlan data as \model{}.}
\begin{tabular}{lcc}
\toprule
Training source (translated CARLA $\to$) & Aux.\ control & mIoU \\
\midrule
Raw CARLA (no translation)             & ---            & 23.2 \\
Seg-conditioned                        & semantic map   & 24.6 \\
Depth-conditioned                      & depth map      & 27.5 \\
\model{} (Ours), resolution-matched         & \textbf{none}  & 28.2 \\
Edge$^\dagger$-conditioned (nuPlan-finetuned) & edge map       & 29.0 \\
Edge-conditioned (off-the-shelf)       & edge map       & 29.6 \\
\textbf{\model{} (Ours), full (hi-res, temporal)} & \textbf{none} & \textbf{31.3} \\
\bottomrule
\end{tabular}
\label{tab:downstream}
\end{table}

We highlight three findings. \textbf{(1) Translation produces large, real downstream gains.} A segmenter trained on \model{}-translated frames improves real-domain mIoU by $+8.1$ over raw CARLA ($23.2 \to 31.3$), showing that our output carries concrete, transferable perception value---not merely visual appeal. \textbf{(2) \model{} gives the best downstream transfer among all conditioning strategies, and is the only one requiring no task-specific control input.} Every baseline must extract a per-frame control signal (edges/semantics/depth); \model{} conditions solely on self-supervised DINO features---a single, unified signal---yet ranks first, $+1.7$ over the strongest baseline (off-the-shelf Edge). Both Edge variants score very closely (29.0 vs.\ 29.6), so the ranking is not driven by the Edge fine-tuning detail. \textbf{(3) The advantage is consistent and not an artifact of resolution.} Under a strictly resolution-matched comparison, \model{} already exceeds Seg and Depth and is near the same-data Edge$^\dagger$ (28.2 vs.\ 29.0); the full high-resolution temporal model---itself a core contribution of this paper---then moves clearly to the top (31.3).

We further note that explicit Seg-conditioning is the \emph{weakest} translation (24.6, barely above raw CARLA): preserving the semantic layout alone does not improve transfer---appearance realism does, which is exactly what DINO-feature conditioning provides.

\paragraph{Why this setup measures sim-to-real ability} Training only on synthetic (or synthetic-translated) data and evaluating on real images exposes precisely the domain shift that Sim-to-Real seeks to close; this is the established instrument behind the GTA5/SYNTHIA$\to$Cityscapes benchmarks~\cite{richter2016gta5,ros2016synthia,cordts2016cityscapes} and behind translation-based domain adaptation such as CyCADA~\cite{hoffman2018cycada}, which makes synthetic frames look real, trains on them, and observes real-test accuracy rise in proportion to how well the appearance gap is bridged. Hence, under a fixed downstream model and fixed labels, a higher real-Cityscapes mIoU directly indicates that the translated frames are perceptually closer to the real domain. Using downstream task performance to quantify the quality of generated data is itself an established evaluation paradigm~\cite{ravuri2019cas,kar2019metasim}.

\paragraph{Scope} This is a deliberately lightweight setup---a single small segmenter, a subset of frames, source-only training---so the absolute mIoU is modest by design and the relative ranking, not the absolute number, is the result. In practice, synthetic data is used to \emph{augment} real data via mixed real+synthetic training or unsupervised domain adaptation~\cite{johnsonroberson2017matrix,hoffman2018cycada,hoyer2022daformer}, which yields substantially higher absolute accuracy; we expect the gains shown above to carry over and compound in that regime.

\section{On the Choice of Fidelity Metrics}\label{sec:fvd_note}

We deliberately report sKID and sFID rather than FVD. With only 149 clips per condition, the high-dimensional I3D Fr\'echet covariance estimate underlying FVD is highly sample-sensitive, so a single FVD value can add estimator noise rather than evidence. We therefore rely on sKID---an unbiased kernel-based metric well-suited to small sample sizes~\cite{binkowski2018kid}---together with sFID and the temporal metrics (Motion-S, WarpSSIM) for a robust overall assessment.

\section{Runtime Analysis}\label{sec:runtime}

The additional inference cost of our conditioning pipeline is small. Under 8-GPU inference, the DINOv3 $\times 4$ feature-extraction branch costs 3.9\,s per 93-frame clip, which is below the Cosmos VAE encoding itself (5.3\,s). PCA projection and causal temporal aggregation are negligible in comparison, and the DiT denoising process is unchanged. \model{} therefore adds no meaningful overhead to the overall generation pipeline while replacing the multi-branch control signal extraction (e.g., per-frame depth or segmentation estimation) required by conventional approaches.

\section{VFM Backbone Feature Comparison}

\begin{figure*}[!t]
    \centering
    \includegraphics[width=0.8\linewidth]{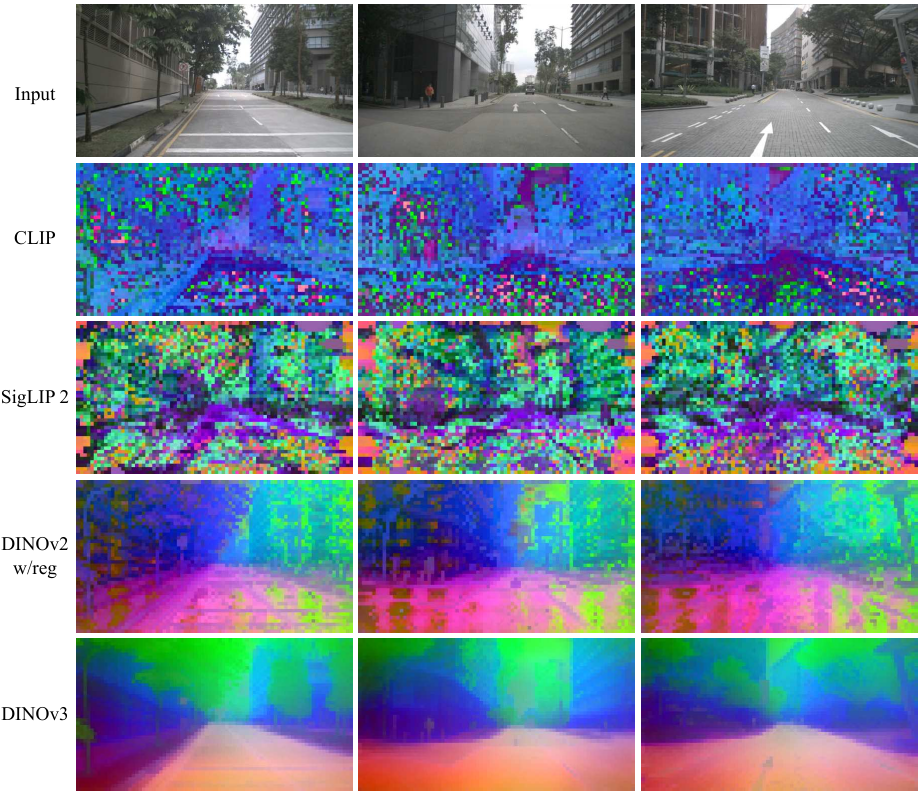}
    \caption{Comparison of dense features. We compare several vision backbones by projecting their dense outputs using PCA and mapping them to RGB. From top to down: CLIP ViT-L/14, SigLIP~2 ViT-g/16, DINOv2 ViT-g/14 with registers, DINOv3 ViT-L/16. Images are forwarded at resolution $1280\times704$ for models using patch size 16 and $1120\times616$ for patch size 14, i.e.\ all feature maps have size $80\times44$.}\label{fig:pca_comp_vfm}
\end{figure*}

Fig.~\ref{fig:pca_comp_vfm} compares dense features extracted by four vision backbones via PCA visualization. Among all backbones, DINOv3~\cite{simeoni2025dinov3} exhibits the highest semantic coherence: regions belonging to the same category (e.g., road, building, vegetation) are mapped to consistent colors with smooth transitions and minimal noise. CLIP~\cite{radford2021learning} and SigLIP~2~\cite{tschannen2025siglip}, by contrast, produce features with coarse spatial resolution and noisy boundaries, reflecting their contrastive pre-training objective that prioritizes global alignment over local structure. DINOv2~\cite{oquab2023dinov2} with registers improves spatial structure over CLIP/SigLIP but still exhibits visible noise and less coherent intra-category representations than DINOv3.

Since PCA introduces sign ambiguity across three components (eight sign combinations) and the mapping from principal components to RGB channels admits six permutations, we enumerate all 48 variants for each backbone and select the visually most informative one for display.

These visual differences directly explain the large performance gap in the DINOv2 vs.\ DINOv3 ablation (Tab.~3 in the main paper, sKID: 22.35 vs.\ 9.08). The noisier DINOv2 features yield less clean inputs for Minor Components Pruning, and its fixed positional embeddings degrade under the high-resolution inputs required by our pipeline, whereas DINOv3's RoPE-box jittering maintains robust encoding at arbitrary resolutions.

\begin{figure*}[!t]
    \centering
    \includegraphics[width=0.9\linewidth]{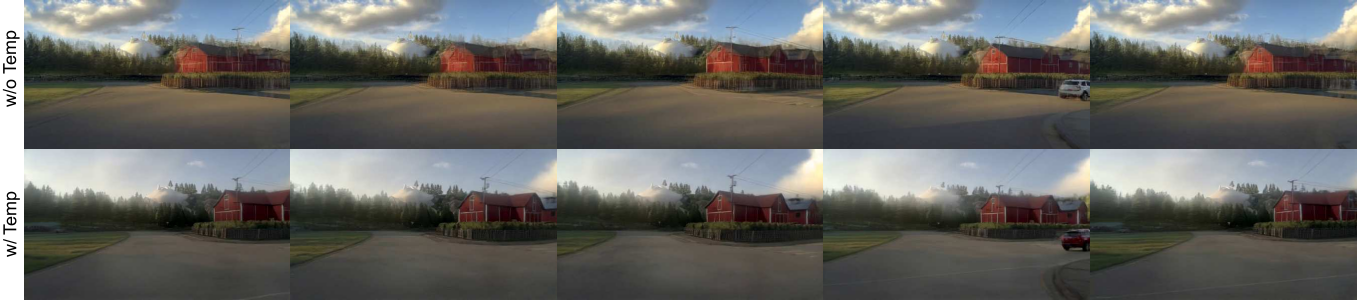}
    \caption{With vs.\ without \textit{Causal Temporal Aggregator}. Top: naive keyframe sampling causes motion blur. Bottom: our aggregator preserves coherence.}\label{fig:appendix_temp_comp}
\end{figure*}

\section{Qualitative Comparisons across Ablated Hyperparameters}\label{sec:qualitative_ablation}

This section provides visual evidence for the ablation results in the main paper.

\subsection{Naive DINO Injection}

As shown in the first row of Fig.~\ref{fig:pca_dim_vis}, directly using non-pruned DINO latents causes severe \textbf{texture leakage}: high-frequency CG surface details (e.g., flat shading artifacts) are faithfully preserved, producing a filtered version of the input rather than a photorealistic regeneration. With Minor Components Pruning, \model{} successfully disentangles structure from texture, preserving geometric layout while enabling realistic rendering.

\subsection{Impact of Minor Components Pruning Degree ($k$)}

\begin{figure*}[!t]
    \centering
    \begin{minipage}[t]{0.42\linewidth}
        \vspace{0pt}
        \centering
        \includegraphics[width=\linewidth]{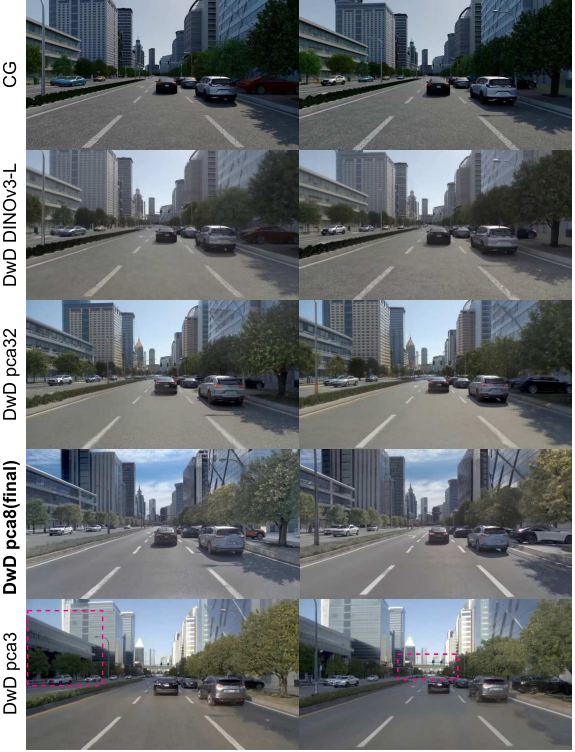}
        \caption{Qualitative comparison across pruning degree ($k$). Naive injection (Row~1): texture leakage. $k{=}3$ (Row~5): structural hallucinations.}\label{fig:pca_dim_vis}
    \end{minipage}\hfill
    \begin{minipage}[t]{0.56\linewidth}
        \vspace{20pt}
        \centering
        \includegraphics[width=\linewidth]{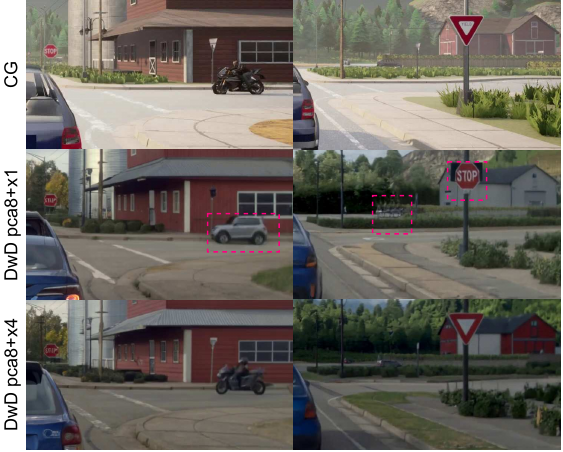}
        \caption{Comparison across upscaling factors $\times S$. \textcolor{red}{Red boxes}: inconsistencies w.r.t.\ CG input.}\label{fig:high_resolution_comp}
    \end{minipage}
\end{figure*}

The PCA truncation factor $k$ governs the trade-off between structural guidance and generative flexibility (Fig.~\ref{fig:pca_dim_vis}):
\begin{itemize}[leftmargin=*, nosep]
    \item \textbf{Insufficient pruning ($k{=}32$):} Excessive feature variance is retained, over-constraining the model with domain-specific noise and degrading realism.
    \item \textbf{Excessive pruning ($k{=}3$):} The control signal becomes too sparse for complex geometries, leading to structural hallucinations. For instance, in Fig.~\ref{fig:pca_dim_vis} (Row~5), a bridge is erroneously synthesized in the distant background.
\end{itemize}

\subsection{Effect of Spatial Upscaling Factor ($S$)}

Fig.~\ref{fig:high_resolution_comp} reveals a positive correlation between $S$ and structural consistency. At $S{=}4$, fine-grained geometric cues enable precise alignment of generated instances with the CG layout. At $S{=}1$, however, the model lacks sufficient spatial detail to disambiguate similar structures---e.g., street lamps are mistranslated into traffic lights (\textcolor{red}{red boxes}).

\subsection{Efficacy of the Causal Temporal Aggregator}

Fig.~\ref{fig:appendix_temp_comp} compares our Causal Temporal Aggregator against naive keyframe sampling. Under fast camera jittering, the naive approach produces visible motion blur and temporal discontinuities (top row). Our aggregator preserves historical motion context through causal convolutions, yielding sequences with superior temporal coherence (bottom row).

\begin{figure*}[!t]
    \centering
    \includegraphics[width=\linewidth]{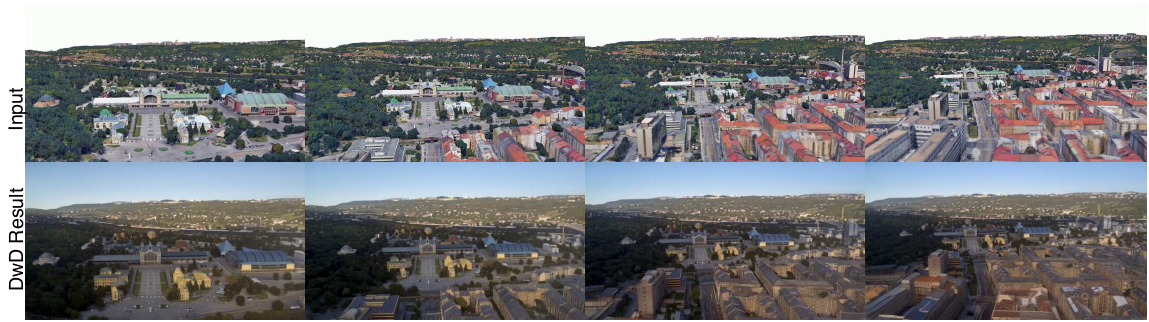}
    \caption{Out-of-domain generalization to aerial urban scenes rendered from a photogrammetric 3D mesh. \model{} is applied without any domain-specific fine-tuning. Prompt: ``Aerial view of a European city on a sunny afternoon, photorealistic, high resolution.''}\label{fig:generalization}
\end{figure*}

\section{Out-of-Domain Generalization}

To probe the generalization capability of \model{} beyond automotive simulation, we apply it to aerial oblique imagery rendered from a photogrammetric 3D surface mesh---a domain entirely unseen during training. As shown in Fig.~\ref{fig:generalization}, despite being trained exclusively on street-level nuPlan~\cite{caesar2021nuplan} driving videos, \model{} produces visually coherent results on aerial urban scenes: building facades receive more natural lighting and material appearance, vegetation gains realistic color variation, and the overall atmosphere becomes more photorealistic.

We attribute this out-of-domain generalization to two factors: (1)~the domain-invariant nature of DINOv3~\cite{simeoni2025dinov3} features, which extract meaningful structural representations regardless of viewpoint or scene type; and (2)~the strong generative prior of the Cosmos Predict 2.5~\cite{ali2025world} backbone, pre-trained on diverse video data. Together, these components allow the conditioning pipeline to transfer structural cues from an unseen domain while the diffusion backbone fills in plausible visual details.

Some limitations remain: fine-grained rooftop details are altered and the color palette shifts toward the driving-domain training distribution. Nonetheless, the result suggests that VFM-Prism conditioning generalizes beyond its training domain, opening potential applications in urban planning visualization, satellite-to-realistic rendering, and simulation for aerial robotics.

\section{Robust Long Video Generation using \model{}}

\begin{figure*}[!t]
    \centering
    \includegraphics[width=\linewidth]{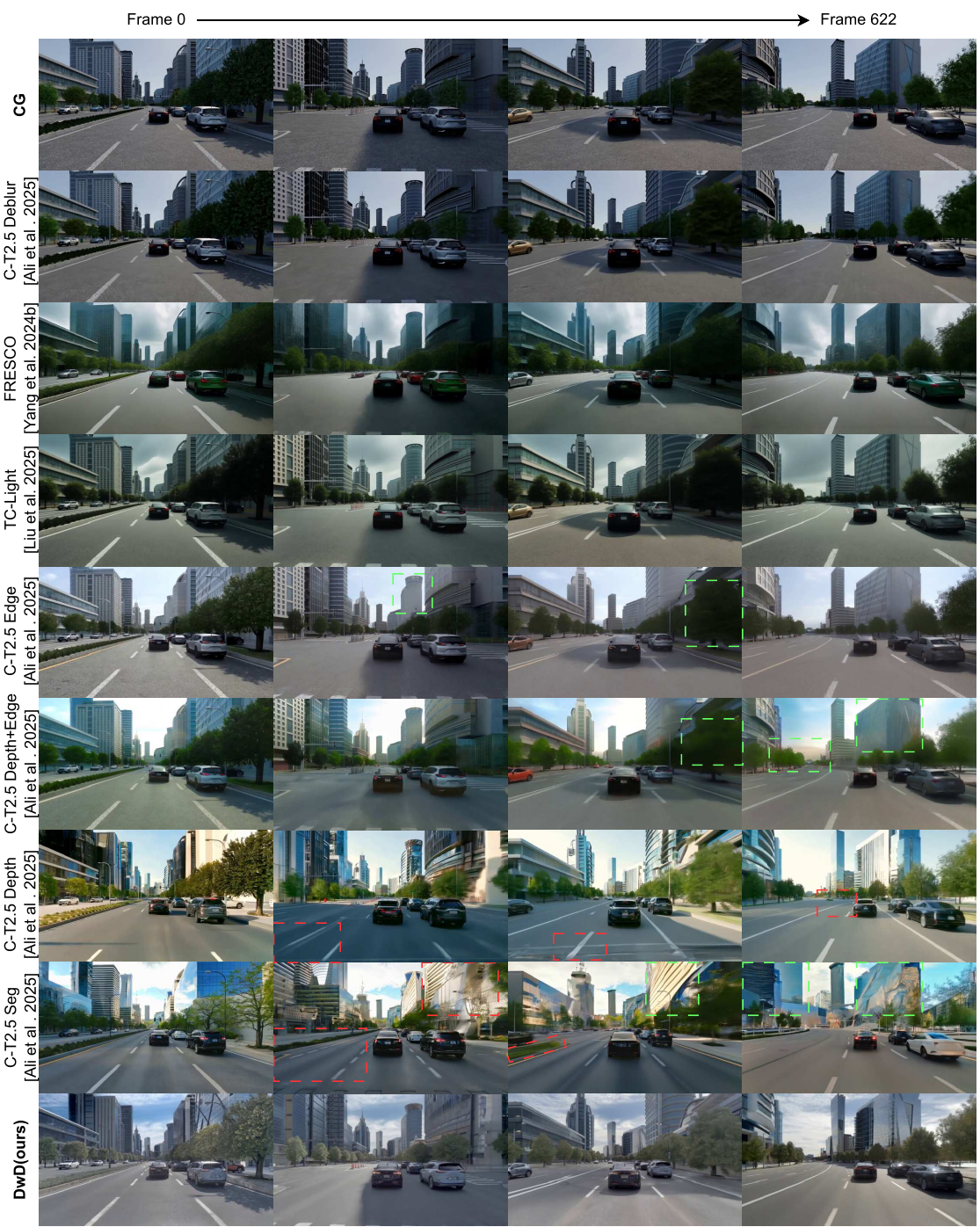}
    \caption{Qualitative comparison of autoregressive long video generation over seven cycles (93 frames each, 651 frames total). Our method maintains visual fidelity throughout, while baseline methods suffer from error accumulation and color oversaturation after approximately three cycles.}
    \label{fig:longvideo}
\end{figure*}

\model{} maintains visual fidelity across extended autoregressive horizons, whereas most Cosmos Transfer 2.5 control branches suffer from severe error accumulation. As shown in Fig.~\ref{fig:longvideo}, each sequence is generated over seven autoregressive cycles with a 93-frame chunk size (651 frames in total). Baseline methods already exhibit color oversaturation and progressive quality degradation after approximately three cycles. \model{} avoids this failure mode because VFM-Prism conditioning provides a semantically stable anchor at each chunk boundary, correcting accumulated pixel-space drift before it compounds. We recommend the supplementary videos for a dynamic evaluation.

\section{Extra Qualitative Comparisons}

\begin{figure*}[!t]
    \centering
    \includegraphics[width=\linewidth]{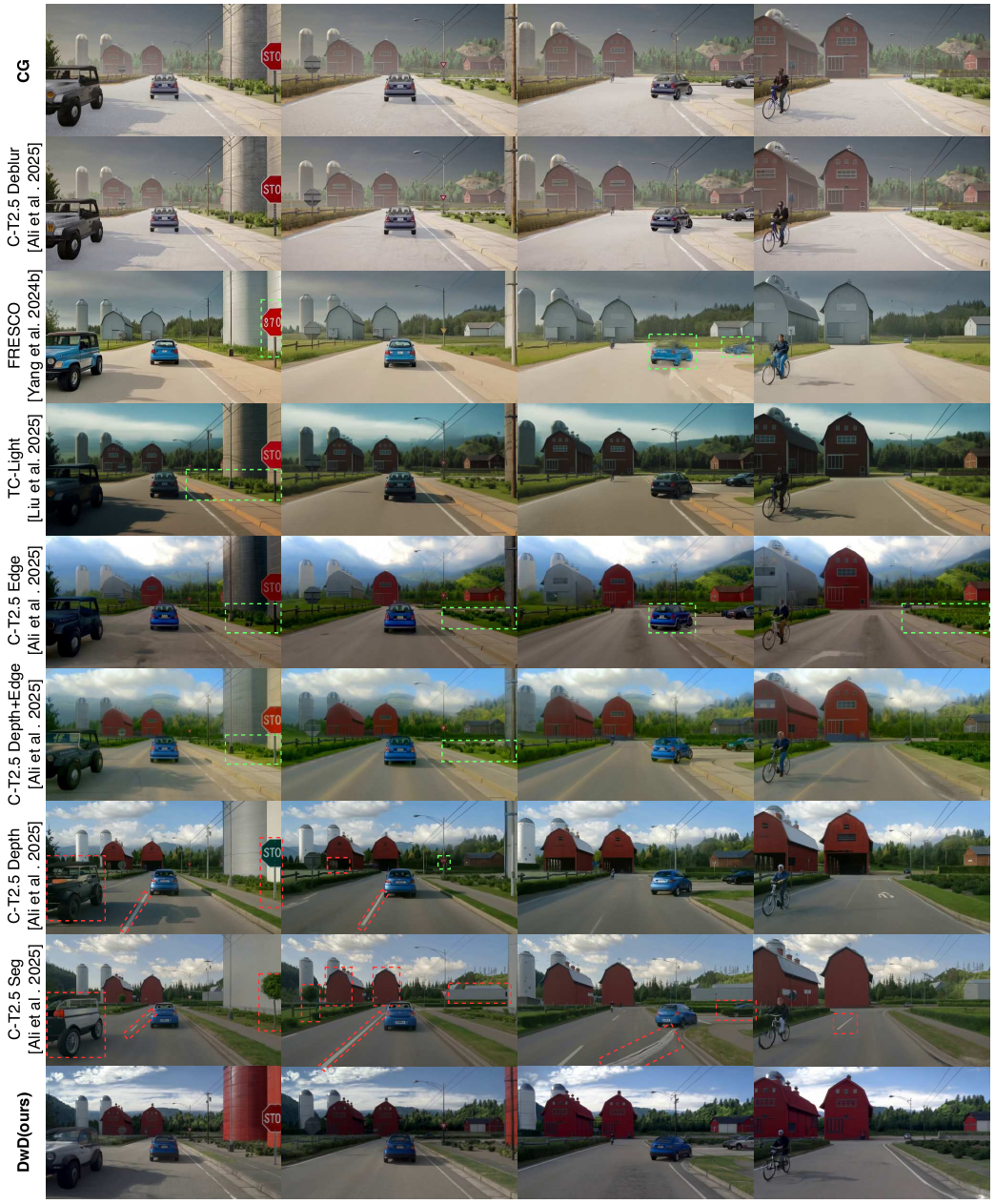}
    \caption{Extra qualitative comparison with state-of-the-art methods. The \textcolor{red}{red boxes} highlight \textcolor{red}{inconsistencies} with respect to the CG input, while the \textcolor{green}{green boxes} point out \textcolor{green}{low-fidelity textures}.}\label{fig:Qualitative_Comparison_extra1}
\end{figure*}

\begin{figure*}[!t]
    \centering
    \includegraphics[width=\linewidth]{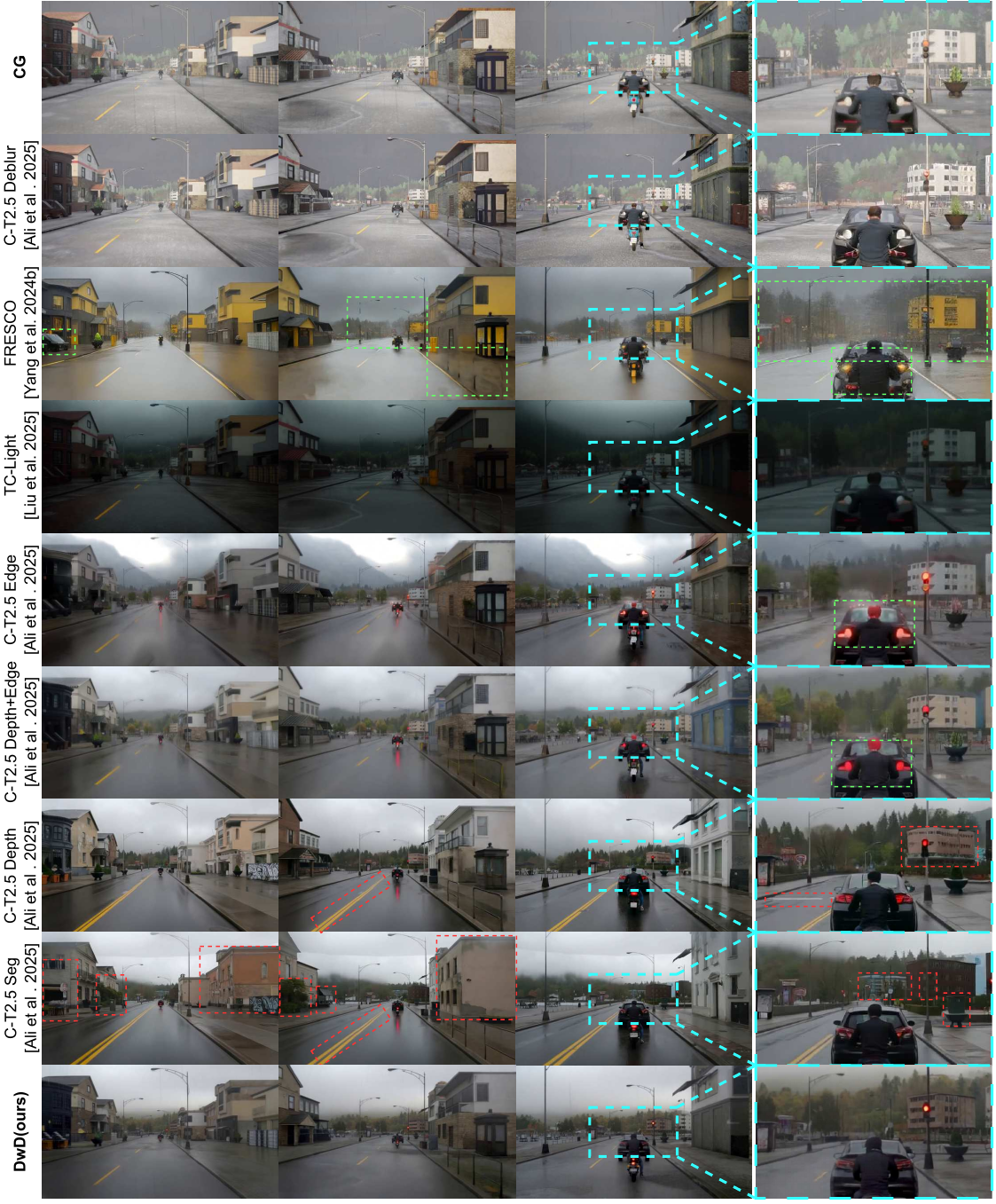}
    \caption{Extra qualitative comparison with state-of-the-art methods. The \textcolor{red}{red boxes} highlight \textcolor{red}{inconsistencies} with respect to the CG input, while the \textcolor{green}{green boxes} point out \textcolor{green}{low-fidelity textures}.}\label{fig:Qualitative_Comparison_extra2}
\end{figure*}

Figs.~\ref{fig:Qualitative_Comparison_extra1} and~\ref{fig:Qualitative_Comparison_extra2} provide additional qualitative comparisons across diverse driving scenes, complementing the main paper's analysis. The same Consistency-Realism Dilemma patterns persist: high-level conditions (Depth, Seg) hallucinate structural elements in ambiguous regions (\textcolor{red}{red boxes}), such as phantom lane markings or misplaced road boundaries, while low-level conditions (Edge, Blur) preserve CG textures almost verbatim (\textcolor{green}{green boxes}), retaining flat shading and synthetic surface patterns. The multi-control variant (Edge+Depth) partially improves realism over Edge alone but still exhibits residual CG artifacts.

In contrast, \model{} consistently achieves both photorealistic textures and faithful structural alignment across all scenes, confirming the generalizability of VFM-Prism conditioning beyond the single scene shown in the main paper. Additional video comparisons are available in the supplementary \texttt{dwd\_comparison.html}.

